\documentclass[lettersize,journal]{IEEEtran}
\usepackage{amsmath,amsfonts}
\usepackage{algorithmic}
\usepackage{algorithm}
\usepackage{array}
\usepackage[caption=false,font=normalsize,labelfont=sf,textfont=sf]{subfig}
\usepackage{textcomp}
\usepackage{stfloats}
\usepackage{url}
\usepackage{verbatim}
\usepackage{graphicx}
\usepackage{cite}
\usepackage{framed,multirow}
\usepackage{tikz}
\usepackage{booktabs}
\usepackage{colortbl}

\usepackage[T1]{fontenc}
\usepackage{amsthm,amsmath,amssymb}
\usepackage{mathrsfs}

\newcommand{\tang}[1]{\textcolor{red}{{}#1}}
\newcommand{\he}[1]{\textcolor{blue}{{}#1}}

\begin{document}

\title{UASTrack: A Unified Adaptive Selection Framework for Customized Multi-modal Object Tracking}

\author{He Wang, Tianyang Xu,~\IEEEmembership{Member,~IEEE}, Zhangyong Tang, Xiao-Jun Wu,  Josef Kittler, ~\IEEEmembership{Life Member,~IEEE}
\thanks{

He Wang, Tianyang Xu, Zhangyong Tang, Shaochuan Zhao, and Xiao-Jun Wu (Corresponding author) are with the School of Artificial Intelligence and Computer Science, Jiangnan University, Wuxi 214122, China (e-mail: 7243115005@stu.jiangnan.edu.cn; tianyang.xu@jiangnan.edu.cn; zhangyong\_tang\_jnu@163.com; wu\_xiaojun@jiangnan.edu.cn).

Josef Kittler is with the Centre for Vision, Speech and Signal Processing, University of Surrey, GU2 7XH Guildford, U.K. (e-mail: j.kittler@surrey.ac.uk).}}

\markboth{Journal of \LaTeX\ Class Files,~Vol.~14, No.~8, August~2021}%
{Shell \MakeLowercase{\textit{et al.}}: A Sample Article Using IEEEtran.cls for IEEE Journals}

\IEEEpubid{
}
\maketitle

\begin{abstract}
Multi-modal object tracking has recently gained significant attention, as auxiliary sensors (event, depth, and thermal cameras) provide unique imaging capabilities that complement RGB sensors. 
Most existing trackers are independently designed to handle multi-modal tracking tasks (RGB-T, RGB-E, and RGB-D) as single-task models, making it inherently challenging to leverage the benefits of broader multi-modal training and limiting cross-modal knowledge sharing.
Although some efforts try to develop a unified system capable of addressing all tasks simultaneously, these systems often rely heavily on task priors, manually switching among branches based on task types, limiting the adaptive ability of the multi-modal tracker to function effectively with any pair of modalities.
To address these limitations, we propose UASTrack, a unified adaptive selection framework that not only provides a unified architecture but also employs unified parameters to effectively integrate various multi-modal tracking tasks. 
Specifically, we design a modality-aware classifier to achieve modality-adaptive control of processing joint image pairs, eliminating the need for task-specific priors during both training and inference. 
Furthermore, to account for the significant differences in input modality characteristics, we introduce a customized processing strategy tailored to the unique properties of each modality.
This approach enables multiple tasks to adapt effectively to the pre-trained network space simultaneously.
Extensive experiments on RGB-T, RGB-E, and RGB-D tracking benchmarks demonstrate that our UASTrack achieves impressive performance, while requiring only 2.10M of additional training parameters and 1.95G FLOPs. 
The code will be made available at https://github.com/ouha1998/Switcher.
\end{abstract}


\section{Introduction}

\IEEEPARstart{V}ISUAL object tracking \cite{kristan2023first, xu2020accelerated} is a crucial research area in computer vision, focusing on estimating the position and size of an object throughout a video sequence, starting from its initial state in the first frame of the video.
Recent advances highlight the limitations of relying solely on an RGB sensor, leading to increased interest in utilizing auxiliary modalities \cite{zhang2022watch}, such as thermal (T) \cite{mctrack}, event (E) \cite{tenet}, and depth (D) \cite{rgbd1k} to aid tracking.
This shift raises a new research topic in multi-modal object tracking, combining the complementary characteristics of the RGB modality with various auxiliary modalities \cite{vlctrack, KSTrack, TGTrack}.
For example, RGB data is highly sensitive to lighting variations, while thermal data remains stable, facilitating robust tracking even under challenging illumination conditions.
RGB-D tracking utilizes the geometric information provided by depth modality to enhance tracking accuracy.
RGB-E tracking capitalizes on the superior temporal resolution and wide dynamic range of event-based data, enabling precise object tracking even in scenarios involving rapid motion.
These complementary features of RGB and auxiliary modalities emphasize the strengths of distinct multi-modal characteristics in overcoming the limitations of single-modality systems.

\begin{figure}[t]
\centering
\includegraphics[width=0.5\textwidth]{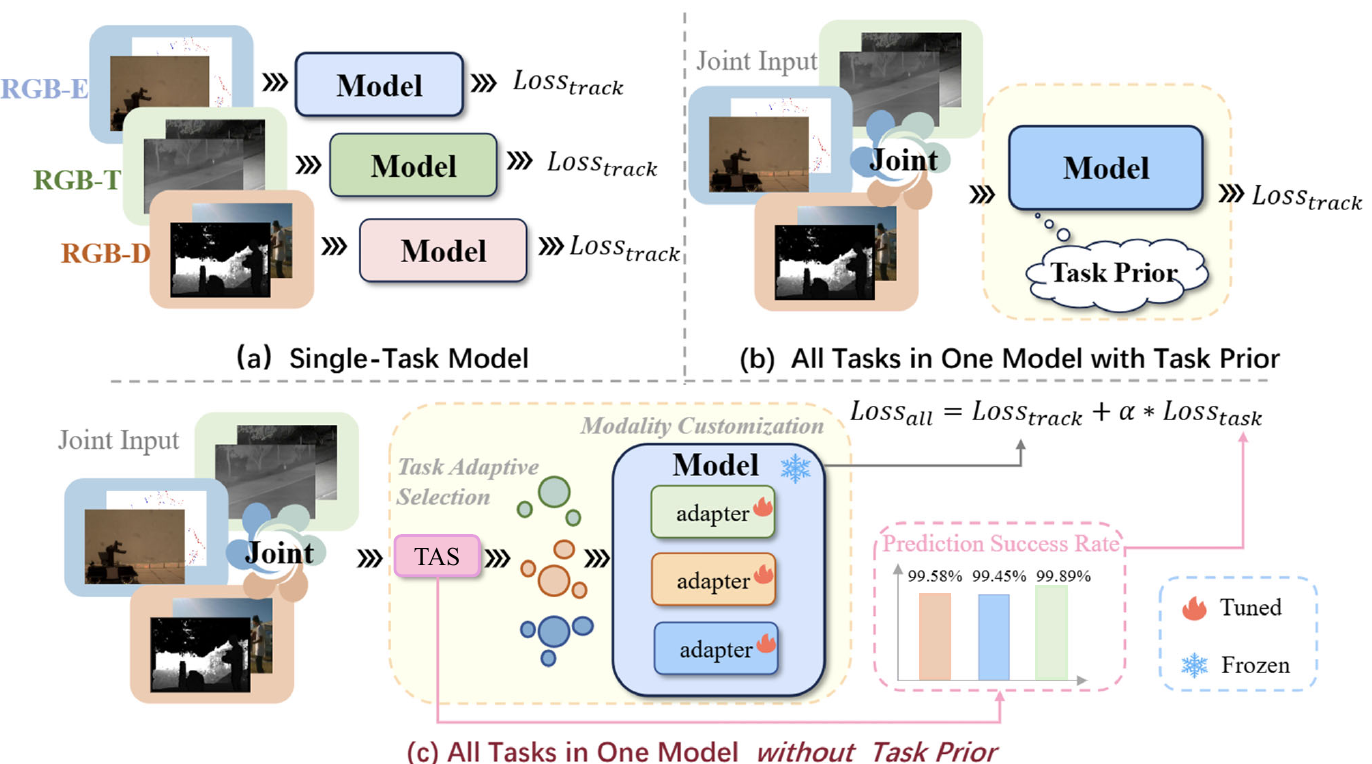}
\caption{
Comparison between the traditional multi-modal object tracking models and our proposed method.
(a) Single-Task models need $N$ task-specific training to cater for $N$ tasks. (b) Accommodating $N$ tasks by one model requires $N$ task priors. (c) Our proposed method, UASTrack, is a unified model with task adaptive selection and modality customization. 
UASTrack uses a single model architecture and a single set of trainable parameters to dynamically accommodate any modality.
It automatically distinguishes the type of modality inputs and applies modality-specific processing tailored to their unique characteristics.
The metric "Prediction Success Rate" quantifies the tracker's capability to dynamically adjust to modality variations, while maintaining robust recognition performance.
}
\label{fig0}
\end{figure}

{Although integrating image pairs formed by combining auxiliary and RGB modalities has gained significant attention in multi-modal object tracking, these approaches still struggle with substantial cross-modal information discrepancies.
To be specific, recent advanced methods that leverage task-specific training strategies to complement the RGB modality \cite{tbsi, mat}, they require $N$ sets of parameters and separate models for each of the $N$ tasks, as shown in Fig. \ref{fig0} (a).
This separate training mechanism leads to redundant training processes, especially in practical applications.
Consequently, there has been growing research interest in developing a unified system for multi-modal object tracking which can efficiently utilize various multi-modal pairs, regardless of their type.}

Constructing a unified multi-modal tracking framework offers several advantages: Firstly, integrating all tasks into one model alleviates the burden of model fine-tuning and hyper-parameter fine-tuning for each individual task, facilitating straightforward comparisons of tracker performance across different modalities. 
Secondly, it enables an effective integration of shared information across all auxiliary modalities.
Currently, there are several approaches that attempt to achieve unification across various multi-modal object tracking tasks, which can generally be classified into two categories.
The first category focuses on employing a unified architecture for multiple tasks. 
For instance, ProTrack \cite{protrack}, ViPT \cite{VIPT}, SDSTrack \cite{sdstrack}, and OneTracker \cite{onetracker} leverage the prompt-tuning paradigm to achieve a unified architecture. 
However, they still require $N$ independent training processes for $N$ tasks, as shown in Fig. \ref{fig0} (a).
The second category method \cite{untrack} addresses the limitations of the first by utilizing a single set of training parameters for all tasks in one model, it relies on manually selecting task priors for video input, as presented in Fig. \ref{fig0} (b).
The task priors specifically refer to the modality labels. 
{Although the SUTrack \cite{sutrack} tries to achieve unified tracker without task labels, it needs a lot of computational cost.
Although SUTrack \cite{sutrack} aims to achieve a unified tracker without using task priors, its high computational cost is a drawback, and the varying adaptability of auxiliary modalities like thermal, event, and depth to RGB-based pre-trained networks further complicates the process.
This highlights the pressing need for a unified framework that effectively leverages complementary information regardless of its origin.}

We aim to build a unified adaptive selection framework with modality-customization for multi-modal object tracking (UASTrack) to address these challenges. 
Our UASTrack is endowed with the ability to recognise the type of auxiliary modality, and incorporates  modality-specific structures reflecting the characteristics of different multi-modal tracking tasks, as shown in Fig. \ref{fig0} (c). 
Specifically, we design a Discriminative Auto-Switch (DAS) dynamically to identify the input tasks by employing a classification mechanism that distinguishes image pairs (e.g., RGB-T, RGB-D, or RGB-E).
The DAS module provides a robust foundation for the subsequent adaptive processing of multi-modal feature fusion.
To enhance the DAS learning capability, we also incorporate Classification Loss (CL) by using cross-entropy.
As illustrated in Fig. \ref{fig0} (c), our proposed DAS module predicts various task identities, achieving a Prediction Success Rate of 99.89\%, 99.58\%, and 99.45\% for the RGB-T, RGB-D, and RGB-E tracking tasks, respectively.
In contrast to previous methods, the DAS functionality enables modality-adaptive tracking.

Although a direct application of an RGB-based pre-trained network has proven to be effective in extracting robust multi-modal data \cite{dfat, VIPT, onetracker}, transforming the features of other modalities into an RGB-based feature space often leads to sub-optimal performance due to the inherent modality gap.
To address this issue, we propose a progressive adapter-learning framework operating at both the feature extraction stage and the task level. 
During feature extraction, bidirectional adapters are inserted within Transformer encoder blocks \cite{transformer} to promote effective interactions between RGB and any auxiliary modality. 
The frozen prediction head that is pre-trained in the RGB domain still exhibits limitations in adapting to fused multi-modal features. 
Different auxiliary modalities possess distinct characteristics: thermal imaging \cite{GMMT} ensures robustness under low-light conditions, event cameras \cite{tenet} offer exceptional temporal resolution for capturing high-speed motion, and depth sensing \cite{rgbd1k} is effective in handling occlusion scenarios. 
Moreover, at the task-level stage, feature maps processed by deeper layers contain more comprehensive information compared to those at the feature extraction stage. 
Therefore, we introduce a modality-customization mechanism at the task level, which further facilitates effective adaptation.

Our approach establishes a unified multi-modal object tracking system, capable of adaptively recognising multi-modal {identities} and integrating {modality-specific refinements }for each task.
Compared to the baseline, which requires 56.44G FLOPs and 92.13M parameters, our proposed UASTrack introduces a modest increase of only 2.10M 
parameters and 1.95G FLOPs, resulting in a significant improvement of 8.5\% in Success Rate on the LasHeR benchmark.

In summary, our contributions are as follows:

\begin{itemize}
   
    \item 
    We propose a unified multi-modal tracker that utilizes a Discriminative Auto-Switch, eliminating the need for prior identification of input auxiliary modality and enabling dynamic modality awareness. Our proposed method establishes a foundation for cross-modal knowledge sharing, reducing redundant training processes across various tracking tasks.
    \item 
    We propose a Task Customization Adapter, enhancing the adaptability of the RGB-based pre-trained model to the multiplicity of modality specific feature spaces and enabling modality-specific customization for different tracking tasks based on their characteristics.
    \item 
    {Extensive evaluations across five benchmarks validate the effectiveness and efficiency of UASTrack, demonstrating a significant improvement over other multi-modal trackers.}
   
\end{itemize}

\section{RELATED WORK}

\subsection{Multi-modal Object Tracking}
Visual object tracking \cite{ostrack, STARK} has wide-ranging applications across various scenarios, such as autonomous driving, mobile robotics, video surveillance, and human-robot interaction. 
The RGB modality \cite{liu2022attention}, as a fundamental data source, provides rich color and texture information, which is essential for object localization.
However, the performance and stability of visual object tracking are limited in complex scenarios involving low illumination, fast motion, and occlusion. 
To address these limitations, multi-modal object tracking \cite{MFGNet,taat, mv-rgbt} incorporating an auxiliary modality (thermal, event, and depth) \cite{cheng2025one,tfnet,zhu2024pr} with the RGB modality exhibits superiority in handling these scenario degradations.
Specifically, depth sensor \cite{rgbd1k} assists in handling the scenario with varying geometric distances, thermal sensor \cite{mpt} enhances the tracker robustness under low illumination, and event sensor with its low-latency motion capture capability (1 \textmu s) \cite{visevent} boosts the high-speed awareness.
Harnessing the unique strengths of RGB and auxiliary modalities, multi-modal information significantly improves the robustness and efficacy of multi-modal object tracking in overcoming complex environmental challenges.

\begin{figure*}[t]
\centering
\includegraphics[width=1\textwidth]{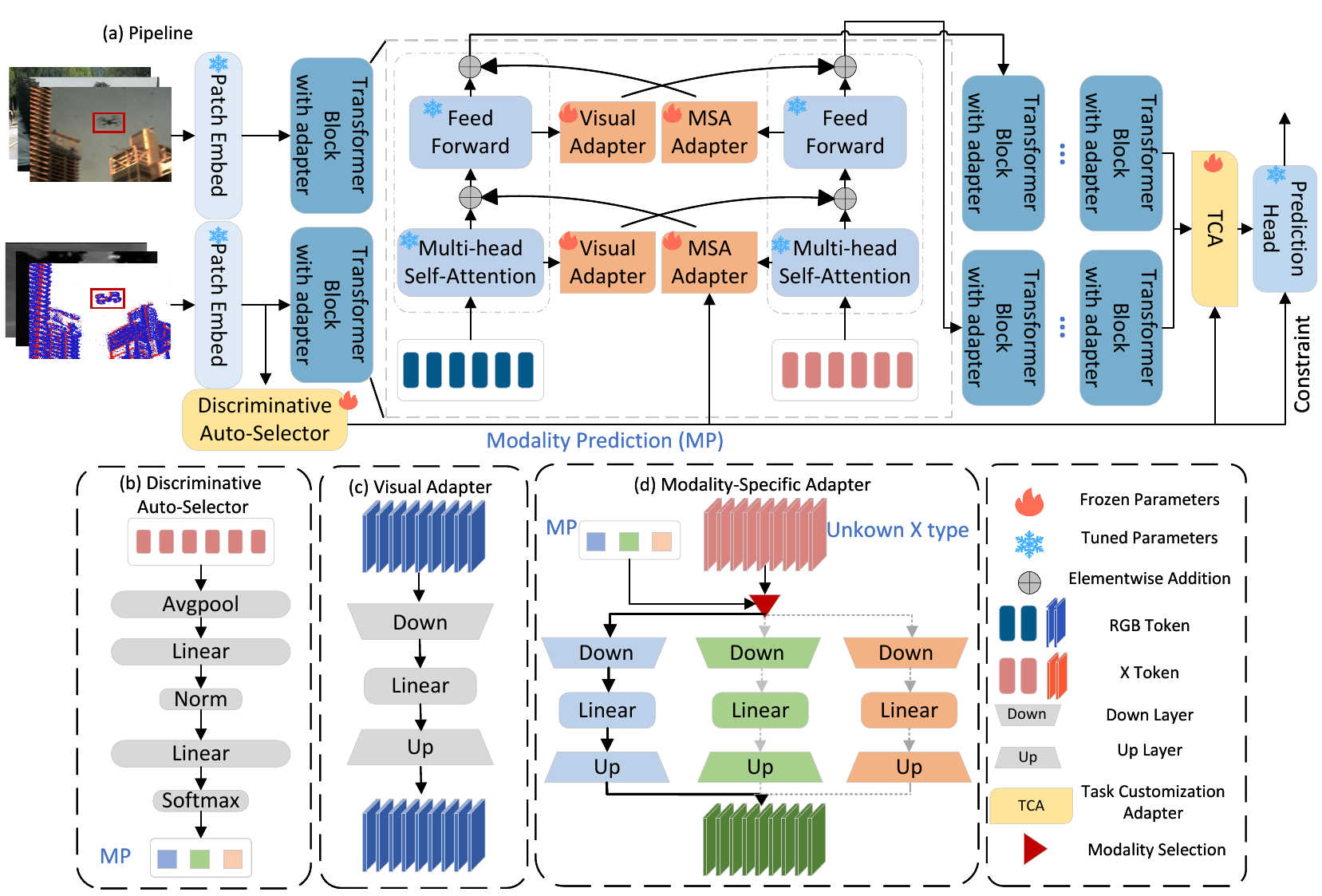}
\caption{An overview of our proposed UASTrack. {In Fig. (d), $\rightarrow$ represents the actual paths adaptively selected by the X features while $\dashrightarrow$ indicates the paths that are not taken.}}
\vspace{-3mm}
\label{fig:2}
\end{figure*}


{RGB-T Tracking: Recently, deep learning-based RGB-T tracking have gained significant attention \cite{transformer}.
These approaches address several significant challenges in RGB-T tracking, such as multi-modal information fusion, feature alignment, and temporal consistency.
To be specific, AETrack \cite{aetrack} and AINet \cite{ainet} excel in emphasizing multi-modal information fusion by integrating complex features from both RGB and thermal data to improve tracking robustness. 
Similarly, multi-modal feature alignment is tackled by methods like VLCTTrack \cite{vlctrack} and CKD \cite{ckd} which align features from different modalities to ensure consistency and better tracking accuracy. 
Additionally, dynamic temporal enhancement is the focus of the STMT \cite{stmt}, which optimizes temporal consistency to improve tracking performance.}



{RGB-E Tracking: In recent years, there has been increasing interest in combining RGB frames with event streams for multi-modal object tracking. 
This integration aims to leverage the complementary strengths of RGB, providing rich visual information, and event streams, offering high-speed temporal data.
For example, CEUTrack \cite{ceutrack} introduces a unified single-stage Transformer framework that integrates color and event modalities for efficient and accurate object tracking.
Building on this, HRCEUTrack \cite{hrceutrack} proposes a mask modeling strategy for encouraging the interaction between the cross-modal tokens and designs an orthogonal high-rank loss for suppressing network fluctuations induced by masking.
HPL \cite{hpl} proposes a novel prior knowledge-driven hybrid prompt learning framework to mitigate the heterogeneity between RGB and event modalities.
}

{RGB-D Tracking: 
A variety of trackers have been proposed to realise effective fusion and interaction between RGB and depth modalities \cite{depthrefiner}. 
For example, DAL \cite{dal} integrates depth information into RGB features based on the correlation filter tracking framework. 
DeT \cite{det} adapts RGB-only tracking architectures by fusing feature maps through pixel-wise operations. 
Similarly, SPT \cite{rgbd1k} utilises separated transformer encoders for RGB and depth, followed by a dedicated fusion module. 
}

{Summary:
Such methods \cite{tfnet,vlctrack,stmt, proformer} require $N$ independent training processes and models for $N$ different tasks, making them computationally inefficient and practically limited.
A unified approach that would accommodate all tasks within an integrated system would provide a more effective solution to handle various tracking scenarios. 
In this work, we propose a unified single-object tracking framework that incorporates a dynamic modality classification system, creating an adaptive tracker for diverse combinations of input modalities. 
Both the architecture and its parameters are shared across all tracking tasks.
Only one training is required to handle multiple task scenarios.}

\subsection{{Unified Multi-modal Trackers}}
Recently, there has been a growing interest in developing unified systems for multi-modal object tracking. 
Several multi-modal trackers, such as ProTrack \cite{protrack}, VIPT \cite{VIPT}, OneTracker, SDSTrack \cite{sdstrack}, {CMDTrack \cite{cmdtrack}}, {STTrack \cite{sttrack}} attempt to establish a unified model for all $N$ tasks. 
But these methods still require $N$ sets of parameters, which restricts their flexibility and cross-modal knowledge sharing.

{Interestingly, Un-Track \cite{untrack} and EMTrack \cite{emtrack} try to use a single set of parameters for all tasks. 
Instead, they rely on modality priors to guide the processing of diverse input types. 
Although M\textsuperscript{3}Track \cite{M3Track} and SUTrack \cite{sutrack} try to build the unified multi-modal tracker without any task priors, it still lacks modality-specific processing due to the significant differences between diverse auxiliary modalities.} 
{In contrast, our proposed UASTrack presents a unified multi-modal tracker to enable modality-specific tracking by introducing a lightweight modality classifier. This is instrumental to achieving excellent performance in diverse multi-modal tracking scenarios by virtue of performing modality-specific customization to adapt to their respective spatial structures.}

\section{Methods}

\subsection{Overall Framework}
The framework we propose for unified task-adaptive multi-modal tracking is illustrated in Fig. \ref{fig:2}. 
The overall architecture consists of a frozen foundation tracker, along with trained components including a Discriminative Auto-Switch (DAS), a Visual Adapter (VA), a Modality-Specific Adapter (MSA), and a Task-Customized Adapter (TCA).
These trained components facilitate task-agnostic representation learning on the inference stage. 
We provide detailed descriptions of the foundation tracker in Section \textit{B}, the task-agnostic representation learning in Section \textit{C}, and the objective loss formulation in Section \textit{D}.

\subsection{The Foundation Tracker} 
As illustrated in Fig. \ref{fig:2}, UASTrack adopts an RGB-based pre-trained Transformer \cite{transformer} as the backbone. 
Multi-modal object tracking aims to predict the bounding box of the target in subsequent frames based on its location and shape in the first frame of a video sequence.
Reliable tracking accuracy requires an effective integration of multi-modal inputs, including RGB images {${{{I}}}_\mathrm{RGB}$ ${\in \mathbb{R}^{\mathrm{H} \times \mathrm{W} \times 3}}$} and auxiliary modality images {${{{I}}_\mathrm{X}}$  ${\in \mathbb{R}^{\mathrm{H} \times \mathrm{W} \times 3}}$} {(X data is commonly represented as three-channel images in current tracking datasets).}
Initially, the foundation network preprocesses input image pairs, converting them into a unified embedding format.
Subsequently, these embedding features are processed by the feature extractor $\mathrm{{F}}$, which generates fused features.
The fused features are then passed to the prediction head $\mathrm{Head}$, which extracts task-relevant information and generates the final prediction location ${\textit{P}}$ through post-processing.
The process of multi-modal object tracking can be described as follows:

\begin{equation}
   \centering \label{equ:1}
{\mathrm{\textit{P}} = \mathrm{Head}(\mathrm{{F}}({{I}}^{\mathrm{RGB}}, {{I}}^\mathrm{X})),}
\end{equation}

{Considering the scarcity of comprehensive multi-modal datasets that simultaneously encompass RGB-T, RGB-D, and RGB-E modalities, as well as the absence of pre-trained multi-modal models, we adopt the RGB-based pre-trained OSTrack \cite{ostrack} as the foundation model to enhance the feature extraction ability of multi-modal tasks. 
Our UASTrack fine-tunes only {2.10M} adapter-learning parameters, which constitute just 2.27\% of the 92.13M parameters in the foundation model, demonstrating preferable transfer learning capabilities.} 

{A Discriminative Auto-Switch is employed to differentiate among various tasks by analyzing the characteristics of the auxiliary modality. 
Our proposed method further facilitates targeted processing of video data by enabling the dynamic selection of the corresponding architecture, thereby achieving modality-specific customization.}

\subsection{Task-Agnostic Representation Learning
}

{
UASTrack integrates lightweight adapters \cite{vitadapter,Madapter} into the foundation model \cite{ostrack}, enabling existing pre-trained foundation trackers to multi-modal features.
In particular, UASTrack embeds adapters in both the feature extraction and the prediction head, facilitating progressive adapter-based learning across different components of the network.}

\begin{figure}[t]
\centering
\includegraphics[width=0.4\textwidth]{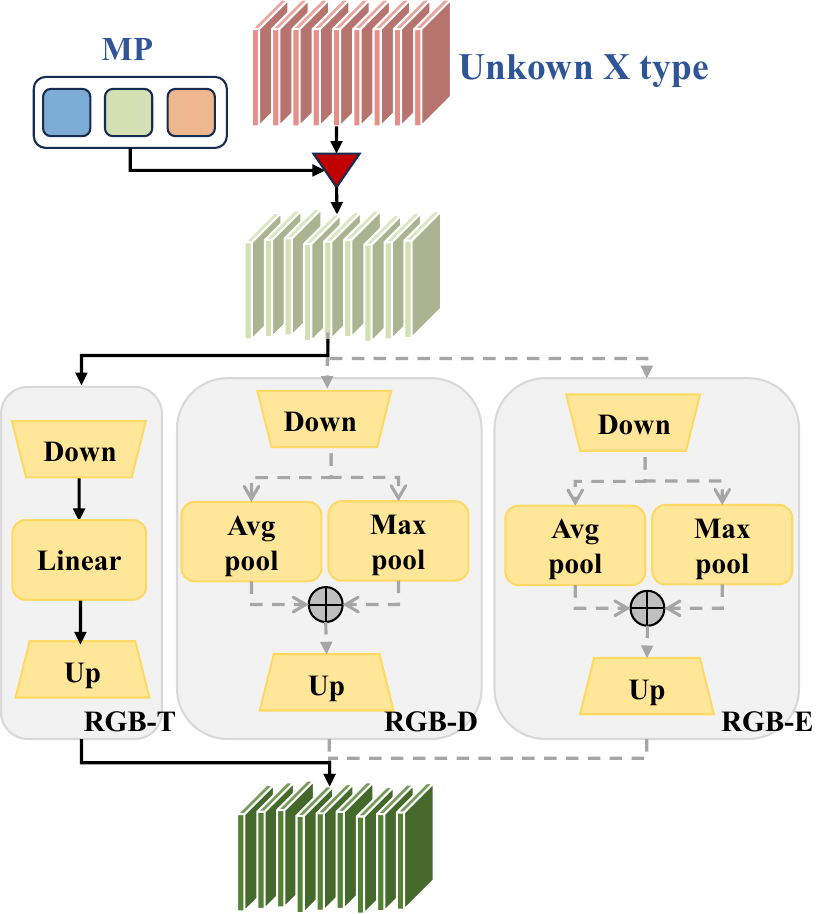}
\caption{Illustration of the proposed Task-Customized Adapter.}
\vspace{-3mm}
\label{fig:3}
\end{figure}

\textbf{Discriminative Auto-Switch.} 
To enable task-agnostic representation learning in multi-modal object tracking scenarios, we propose a Discriminative Auto-Switch (DAS) to acquire a modality prediction (MP), which is activated during inference. 
The depth modality \cite{det} provides spatial information, offering accurate measurements of object distance from the camera.
The event modality \cite{MMHT} excels in high-speed motion tracking due to its high temporal resolution, capturing pixel-level intensity changes at microsecond intervals. 
The thermal modality \cite{lasher} is particularly useful in low-illumination environments, as it detects heat signatures, making it effective for detecting objects in the dark scenarios. 
{Based on the significant differences among auxiliary modalities, the DAS is designed to distinguish features during inference in a lightweight and efficient manner.
Therefore, our framework enables task-agnostic representation learning for multiple multi-modal object tracking tasks.}

As illustrated in Fig. \ref{fig:2} (b), the proposed DAS module is inserted between the patch embedding layer and the first transformer block \cite{transformer}.
{Formally, $\boldsymbol{f}_\mathrm{X}$ represents auxiliary features obtained after the patch embedding layer. 
Initially, $\boldsymbol{f}_\mathrm{X}$ passes through an adaptive average pooling layer ($\mathrm{AdaptiveAvgPool}$), which resizes its spatial dimensions to 1×1:}

\begin{equation}
\centering \label{equ:2}
{\boldsymbol{f}_\mathrm{X}^{'} = \mathrm{AdaptiveAvgPool}(\boldsymbol{f}_\mathrm{X})},
\end{equation}

Subsequently, the reshaped features {$\boldsymbol{f}_\mathrm{X}^{'}$} are processed through two linear layers, {$\mathrm{FC}_{1}$} and {$\mathrm{FC}_{2}$}, to calculate the task-specific predicted probability $P_{m}$. 
This process can be formulated as:

\begin{equation}
\centering \label{equ:3}
{{P}_{m} = \mathrm{FC_{2}}(\mathrm{Norm}(\mathrm{FC_{1}}(\boldsymbol{f}_\mathrm{X}^{'}))),}
\end{equation}

To identify the input modality, the $Argmax$ operation is applied to $P_{m}$, returning the index corresponding to the maximum probability value:

\begin{equation}
\centering \label{equ:4}
{MP = \mathrm{Argmax}(P_\mathrm{m}),}
\end{equation}

Here, $MP$ represents the predicted modality and serves as a crucial input for subsequent multi-modal feature fusion and modality-specific optimization. 
As shown in Fig. \ref{fig0}, the proposed DAS achieves high accuracy, demonstrating its effectiveness in modality prediction. 
Specifically, it attains prediction success rates of 99.58\%, 99.62\%, and 99.96\% for the RGB-T, RGB-E, and RGB-D tasks, respectively. 
Importantly, the DAS enables the prediction of the input task types without requiring prior knowledge of the modality. 

 
{
To strengthen the model’s task differentiation capabilities, a task-recognition training strategy employing Classification Loss (CL) is adopted.
The task loss, {$\mathrm{L}_\mathrm{{task}}$}, is defined as the cross-entropy loss between the predicted probability $P_{m}$ and the ground truth modality labels $T_\mathrm{m}$. 
This loss is formulated as:}

\begin{equation}
\centering \label{equ:5}
{\mathrm{L}_\mathrm{task} = - \sum_{i=1}^\mathrm{N} T_{m, i} \log(P_\mathrm{m}),}
\end{equation}
where $N$ denotes the number of multi-modal object tracking tasks.

\textbf{Modality-Specific Adapter.}
As illustrated in Fig. \ref{fig0} (c), cross-modal interactions between RGB features and auxiliary features are enabled by a bidirectional adapter module.
{Each modality offers unique advantages: RGB images provide rich texture details \cite{apfnet}; thermal imaging \cite{sttrack} ensures robustness under low-light conditions; event cameras \cite{tenet} deliver exceptional temporal resolution for capturing high-speed motion; and depth sensing \cite{zhu2024pr} is effective in handling occlusion scenarios.}
To accommodate the distinct characteristics of different modalities, we design task-specific adapter structures with non-shared parameters for the RGB branch and the auxiliary X branch.
Specifically, the modality-specific data $X$ in the $l-th$ encoder block, denoted as {${\boldsymbol{f}_\mathrm{X}^{l} \in \mathbb{R}^{\mathrm{H} \times \mathrm{W} \times \mathrm{C}}}$}, is split based on the previously predicted modality $MP$. 
Here, $C$ represents the number of feature channels.

The features {$\boldsymbol{f}_\mathrm{X}^{l}$} are first processed by a down-sampling layer {$\mathrm{Down}$} to reduce the feature channel dimension. 
Next, a fully connected layer {$\mathrm{FC}_{3}$} is applied to ensure the consistency of modality-specific features, while minimizing the number of trainable parameters.
Then, the features pass through an up-sampling layer {$\mathrm{Up}$} to restore the original channel dimensions.
The task-specific sub-adapters are mathematically formulated as:

\begin{equation}
\centering \label{equ:6}
{\boldsymbol{f}^{'}_\mathrm{X} = \mathrm{Up}(\mathrm{FC_{3}}(\mathrm{Down}(\boldsymbol{f}^{l}_\mathrm{X}))),}
\end{equation}
where $N$ sub-adapters are designed for $N$ tasks. 
This modular approach ensures that each task benefits from a separate feature transformation, enhancing the overall adaptability of the framework.

\textbf{The Visual Adapter}. 
The RGB modality offers comprehensive color information and high-resolution texture details through its red, green, and blue channels. 
In the Visual Adapter (VA), we extract features from the RGB modality, facilitating interaction with auxiliary modalities through bidirectional adapters. 
As illustrated in Fig. \ref{fig0} (d), the VA employs a down-sampling layer, a fully connected layer, and an up-sampling layer to process the RGB features, as described by:

\begin{equation}
\centering \label{equ:61}
{\boldsymbol{f}^{'}_\mathrm{RGB} = \mathrm{Up}(\mathrm{FC_{4}}(\mathrm{Down}(\boldsymbol{f}^{l}_\mathrm{RGB}))),}
\end{equation}

where {$\boldsymbol{f}^{l}_\mathrm{RGB}$} represents the input RGB features.
The VA maintains the same architecture as the task-specific sub-adapter.

\textbf{Task-Customized Adapter.} To enhance the adaptability of the frozen prediction head to downstream multi-modal data, we introduce task-level adapters with a minimal number of trainable parameters.
Simultaneously, to address the significant variations in auxiliary modalities, we implement modality-specific optimization tailored to the unique characteristics of each modality.

{We first analyze the properties of different auxiliary modalities to determine their appropriate processing strategies. 
Specifically, the thermal modality features are more comprehensive and exhibit minimal redundancy, making them suitable for processing through a general adapter module. 
We adopt the same architecture for depth and event features within the Task-Customized Adapter (TCA), based on their shared representational characteristics. 
Specifically, the event modality responds to regions with abrupt motion or intensity changes, resulting in highly sparse feature maps with strong localized activations \cite{event_sparse}.
Similarly, the depth modality captures meaningful information primarily along the geometric contours, while background regions such as walls and floors often yield sparse or redundant features \cite{depth_sparce}.
This similarity motivates us to adopt the same architectural design.
Inspired by TENet \cite{tenet}, we employ a combination of average pooling and max pooling in our design to focus on local responses while preserving global features, thereby improving the representational capability of event and depth modalities. }

In view of these differences, we design modality-customized adapters for depth and event modalities, as illustrated in Fig. \ref{fig:3}. 
These adapters leverage a dual-pooling mechanism: max pooling is employed to extract high-response features, while average pooling is used to retain global-response features. This combination ensures a balanced and robust feature representation, as described by:

\begin{equation}
\centering \label{equ:7}
{\boldsymbol{f}^{'}_\mathrm{X} = \mathrm{Up}(\mathrm{Avg}(\mathrm{Down}(\boldsymbol{f}_\mathrm{X}))
+\mathrm{Max}(\mathrm{Down}(\boldsymbol{f}_\mathrm{X})),}
\end{equation}
where {$\mathrm{Avg}$} and {$\mathrm{Max}$} denote average pooling and max pooling layers, respectively. 
This design enables targeted processing of depth and event data, effectively addressing their sparsity and redundancy while preserving critical information.

\subsection{The Objective Loss Function}
Following OSTrack \cite{ostrack}, we utilize focal loss as the classification loss {$\mathrm{L}_{\mathrm{cls}}$} and adopt {$\mathrm{L}_{1}$} loss and {$\mathrm{L}_\mathrm{GIoU}$} loss for regression. 
To further enhance the learning capability of the Discriminative Auto-Switch, we introduce an additional classification loss {$\mathrm{L}_\mathrm{task}$} that incorporates cross-entropy loss. 
The overall loss $L$ is defined as:

\begin{equation}
\centering \label{equ:8}
{\mathrm{L} =  \mathrm{L}_\mathrm{cls}+\lambda_{1} \mathrm{L_{1}}+\lambda_{2}\mathrm{L}_\mathrm{GIoU}+ \alpha *\mathrm{L}_\mathrm{task}.}
\end{equation}
where $\lambda_{1}$, $\lambda_{2}$, and $\alpha$ are set as 5, 2, and 0.1, respectively.

\begin{table*}[]
\renewcommand{\arraystretch}{1.4}
\centering
\caption{
A comparison with state-of-the-art methods on LasHeR, DepthTrack, and VisEvent benchmarks. 
"Separated" refers to trackers that handle different tasks using distinct models trained with separate parameter sets.
"Unified-Architecture" indicates trackers that share a unified architecture but still require task-specific training parameters.
"Unified-All" denotes trackers that utilize a single architecture and a single set of training parameters to perform multiple tasks.
The best results are highlighted in \tang{red}, while the second-best are shown in \he{blue}, consistently throughout the table. 
}\vspace{0.3mm}
\scalebox{1.2}{ 
\begin{tabular}{cccccccccccc}
\hline 
\multirow{2}{*}{Type} & \multirow{2}{*}{Method} & \multirow{2}{*}{Venue} & \multicolumn{3}{c}{LasHeR} & \multicolumn{2}{c}{VisEvent} & \multicolumn{3}{c}{DepthTrack} \\ 
 &  &  & SR$\uparrow$ & PR$\uparrow$& NPR$\uparrow$& SR$\uparrow$& PR$\uparrow$& Pr$\uparrow$& Re$\uparrow$& F-score$\uparrow$ \\ \hline \hline
\multirow{18}{*}{Separated} 
&TBSI &CVPR23 &0.556 &0.692 &{0.657}& - & - & - & - & - \\
&LSAR &TCSVT23   &0.385 &0.460 &- & - & - & - & - & - \\
& GMMT & AAAI24 & {0.566} & {0.707} & {0.670} & - & - & - & - & - \\
&ProFormer &TCSVT24 &0.533 &0.674 &0.630 & - & - & - & - & -\\
&QueryTrack &TIP24  &0.520 &0.660&- & - & - & - & - & -\\
& BAT & AAAI24 &{0.563} & {0.702} &-  &-  &-  & - &-  &- \\
&{CKD}  &ACMMM24 &0.581&0.732  &\textcolor{blue}{\textbf{0.693}}& - & - & - & - & -\\
&{AINet-256} &AAAI25 & 0.582 &0.730 &0.690 &- &-&- &- &- \\
&{AETrack} &TCSVT25 &\textcolor{blue}{\textbf{0.596}} &\textcolor{blue}{\textbf{0.747}} &\textcolor{red}{\textbf{0.710}} &-&-&-&-&-    \\
&{DMD} &TIP25 &0.576 &0.726 &0.686 &-&-&-&-&-    \\
&CEUTrack &ARXIV24 &-&-&-&0.531 &0.691&-&-& \\	
&MMHT & NN24& - &- &- &0.551 &0.733&-&-&-\\
&TENeT &NN25 &- &-&- &0.601 &0.765 &-&-&-\\
&{HPL} &TCSVT25 &-&-&- &0.611 &\textcolor{blue}{\textbf{\textbf{0.778}}} &- &-&-\\
&SPT &IJCV24 & -& -& -&- &-&0.527 &0.549 &0.538 \\
& CDAAT & SPL24 &- &- &- &- &- &0.578&0.603&0.590 \\
&TABBTrack &PR24 & & & & & &{0.622}&0.615&0.618\\
&{DepthRefiner} &ICME24 &-& -& -&- &-&0.513 &0.507 &0.510 \\
\hline
\multirow{4}{*}{Unified-} 
& Protrack & ACMMM22 & 0.421 & 0.509 & - & 0.474 & 0.617 & 0.583 & 0.573 & 0.578 \\
& ViPT & CVPR23 & 0.525 & 0.651 & - & 0.589 & 0.756 & 0.561 & 0.581 & 0.571 \\
Architecture& OneTracker & CVPR24 & 0.538 & 0.672 & - & {0.608} & {0.767} & 0.607 & 0.604 & 0.609 \\
& SDSTrack & CVPR24 & 0.531 & 0.665 & 0.631 & 0.597 & 0.767 & 0.619 & 0.609 & 0.614 \\
&{CMDTrack(S12)} &TPAMI25 &0.566 &0.688 &- &\textcolor{blue}{\textbf{0.613}}&0.758 &0.591 &0.607 &0.598 \\
&{STTrack} & AAAI25&\textcolor{red}{\textbf{0.603}} &\textcolor{red}{\textbf{0.760}} &-&\textcolor{red}{\textbf{0.619}} &\textcolor{red}{\textbf{0.786}}&\textcolor{red}{\textbf{0.632}}&\textcolor{red}{\textbf{0.634}}&\textcolor{red}{\textbf{0.633}} \\
\hline
\multirow{5}{*}{Unified-All} & Un-Track & CVPR24 & 0.511 & 0.604 & 0.640 & 0.592 & 0.735 & 0.566 & 0.588 & 0.577 \\
& {EMTrack} & TCSVT24 & 0.533 &0.659 & - & 0.584 &0.724 &0.580 &0.585 &0.583
\\
& {M\textsuperscript{3}Track} & SPL25 & 0.525 & 0.658 & 0.622 &0.596 &0.767 &0.564&0.585&0.574 \\
& {SUTrack-B224*} & {AAAI25} 
&0.563 &0.708 &0.670 &0.601 &0.771 &0.625 &0.622 &0.624 \\
& \textbf{UASTrack} & - & {\textbf{0.570}} & {\textbf{0.711}} &{\textbf{0.675}}  & {\textbf{0.610}} & {\textbf{0.773}} & \textcolor{blue}{\textbf{0.630}} & \textcolor{blue}{\textbf{0.625}} & \textcolor{blue}{\textbf{0.628}} \\ \hline
\end{tabular}}
\label{tab1}
\end{table*}

\section{Experiments}\label{stylefiles}
To evaluate the advantages of our proposed UASTrack, we compare UASTrack performance against both separated training trackers and unified trackers. 
The comparison methods includes {SUTrack \cite{sutrack}}, Un-Track \cite{untrack}, OneTracker \cite{onetracker}, ViPT \cite{VIPT}, SDSTrack \cite{sdstrack}, TBSI \cite{tbsi}, GMMT \cite{GMMT}, BAT \cite{bat}, APFNet \cite{apfnet}, QueryTrack \cite{querytrack}, LSAR \cite{lsar}, ProFormer \cite{proformer}, CAT++ \cite{cat++}, TENeT \cite{tenet}, SPT \cite{zhu2024unimod1k}, ProTrack \cite{protrack}, CEUTrack \cite{ceutrack}, MMHT \cite{MMHT}, TABBTrack \cite{TABBTrack}, CDAAT \cite{CDAAT}, OSTrack \cite{ostrack}, {CMDTrack \cite{cmdtrack}, STTrack \cite{sttrack}, AETrack \cite{aetrack}, AINet \cite{ainet}, CKD \cite{ckd}, DMD \cite{dmd}, HPL \cite{hpl}, M\textsuperscript{3}Track \cite{M3Track}, DepthRefiner \cite{depthrefiner}, and TVTrack \cite{tvtracker}}.
Our foundation network utilizes {OSTrack-256} \cite{ostrack} as the pre-trained model.
{We compare our model with SUTrack-B224 because our model uses OSTrack-256 as the baseline, and their model sizes are similar. To ensure a fair comparison, we use the combination of training data from DepthTrack, VisEvent, and LasHeR to train the SUTrack-B224, which differs from the original SUTrack training that utilizes a broader dataset. }

\begin{table}[]
\renewcommand{\arraystretch}{1.5}
\centering
\caption{A comparison with state-of-the-art methods on other RGB-T tracking datasets including on RGBT234 and GTOT datasets.}\vspace{0.3mm}  
\scalebox{1}{
\begin{tabular}{cccccc}
\hline 
\multirow{2}{*}{Type} & \multirow{2}{*}{Method} & \multicolumn{2}{c}{RGBT234} & \multicolumn{2}{c}{GTOT} \\
 &  & SR$\uparrow$ & PR$\uparrow$ & SR$\uparrow$ & PR$\uparrow$ \\ \hline \hline
\multirow{7}{*}{Separated} & APFNet & 0.579 & 0.827 & 0.739 & 0.905 \\
& TBSI & 0.637 & 0.871 & - & - \\
& BAT &0.641 &0.868 & 0.763 &0.909 \\
& QueryTrack &0.600 &0.841 &0.759 &0.923 \\
& CAT++ &0.592 &0.840 &0.733 &0.915 \\
&{AINet} &\textcolor{blue}{\textbf{0.668}}&0.891&-&-\\
& {CKD}&\textcolor{red}{\textbf{0.674}} &\textcolor{red}{\textbf{0.900}}&\textcolor{blue}{\textbf{0.772}}&\textcolor{blue}{\textbf{0.932}}\\
& {DMD}&0.667 &{0.893}&{0.768}&{0.924}\\
\hline
\multirow{4}{*}{Unified-Architecture}
& Protrack & 0.587 & 0.786 & -&-\\
& ViPT & 0.617 & 0.835 & - & - \\
& SDSTrack & 0.625 & 0.848 & 0.760 & 0.887 \\
&{STTrack} &0.667 &\textcolor{blue}{\textbf{0.898}} &- &- \\
\hline
\multirow{2}{*}{Unified-All} & Un-Track & 0.618 & 0.837 & - & - \\
 & UASTrack & \textbf{0.651}& \textbf{0.876} & \textcolor{red}{\textbf{0.789}} & \textcolor{red}{\textbf{0.933}} \\ \hline
\end{tabular}}
\label{tab2}
\end{table}

\begin{table*}[]
\centering
\caption{An ablation study of our proposed components. The column $\sigma$ represents the average percentage change across all metrics compared to the baseline. Our plain version is highlighted in \textbf{Bold}. {The '+' indicates the additional computational cost added on top of the baseline.}
}\vspace{0.3mm} 
\renewcommand{\arraystretch}{1.3}
\scalebox{1.1}{
\begin{tabular}{ccccccccccccccc}
\hline
& &  &  & & & & \multicolumn{2}{c}{LasHeR} & \multicolumn{2}{c}{VisEvent} & \multicolumn{3}{c}{DepthTrack} \\
\multirow{-2}{*}{Row}&\multirow{-2}{*}{DAS} & \multirow{-2}{*}{CL} & \multirow{-2}{*}{TCA} & \multirow{-2}{*}{{Params}} & \multirow{-2}{*}{{Flops}} &\multirow{-2}{*}{{FPS}} & SR & PR & SR & PR & Pr & Re & F-score 
&\multirow{-2}{*}{$\sigma$} 
\\ \hline \hline
\#1&- & - & - &92.13M &56.44G &71.28 & 0.482 & 0.609 & 0.588 & 0.754 & 0.577 & 0.582 & 0.579 &-\\
\#2&\checkmark & & & +0.59M&+0 & 48.23 & 0.535 & 0.678 & 0.591 & 0.760 & 0.583 & 0.585 & 0.584&+2.07\% \\
\#3&\checkmark &\checkmark & &+0.59M&+0 & 48.17 & 0.551 & 0.687 & 0.602 & 0.766 & 0.603 & 0.611 & 0.609 &+3.68\%  \\
\#4&\checkmark &  &\checkmark & +1.48M&+0.08G & 46.48 & 0.547 & 0.682 & 0.595 & 0.766 & 0.601	&0.593	&0.597 &+3.00\%\\
\#5&\checkmark  &\checkmark   &\checkmark & +1.48M&+0.08G & 43.93&{\textbf{0.570}} & {\textbf{0.711}} &{\textbf{0.610}} & \textbf{0.773} & \textbf{0.630} & \textbf{0.625} & \textbf{0.628} & \textbf{+4.86}\% \\ \hline
\end{tabular}}
\label{tab3}
\end{table*}

\begin{table*}[]
\centering
\caption{Ablation study for our proposed Task-Customized Adapter (TCA).}\vspace{0.3mm}
\renewcommand{\arraystretch}{1.3}
\scalebox{1.15}{
\begin{tabular}{cccccccccc}
\hline
\multirow{2}{*}{Row}&\multirow{2}{*}{Method} & \multicolumn{2}{c}{LasHeR} & \multicolumn{2}{c}{VisEvent} & \multicolumn{3}{c}{DepthTrack} \\
& & SR & PR & SR & PR & Pr & Re & F-score  \\ \hline \hline
\#1&\multicolumn{1}{c}{w/o maxpool} & 0.558 & 0.695 & 0.603 &0.767  & 0.614 & 0.605 & 0.609  \\
\#2&\multicolumn{1}{c}{w/o avgpool} & 0.554 & 0.689  & 0.596 &0.764 & 0.602 & 0.596 & 0.599  \\
\#3&\multicolumn{1}{c}{w/ avgpool+maxpool} &0.556  &0.692  & \textbf{0.610} &\textbf{ 0.773} & \textbf{0.630} & \textbf{0.625} & \textbf{0.628} \\
\#4&\multicolumn{1}{c}{w/ linear} &\textbf{ 0.570} & \textbf{0.711} & 0.602 & 0.766 & 0.606 & 0.611 & 0.609  \\ \hline
\end{tabular}}
\label{tab4}
\end{table*}

To train our proposed UASTrack, only the parameters in Discriminative Auto-Switch (DAS), Modality-Specific Adapter (MSA), and Task-Customized Adapter (TCA) are learnable, as depicted in Fig. \ref{fig:2}. 
In addition to the visual data object tracking loss, we incorporate a cross-entropy loss that enhances DAS's classification ability for various tasks. 
{Our proposed methodology is implemented in PyTorch and executed on a system with an Intel (R) Core (TM) i9-9980XE CPU and an NVIDIA 3090Ti GPU.}

{Regarding the training setup, the entire framework, including the Discriminative Auto-Selector, Visual Adapter, Modality-Specific Adapter, Task Customization Adapter modules, is trained in an end-to-end manner. }

We set the batch size to 32, training for 80 epochs. 
The learning rate for the backbone is set to 4e-4, with a decay ratio of 0.8.
We adopt the AdamW optimizer with a weight decay of 1e-4. 
Additionally, template feature dimensions are uniformly resized to 128×128, while the search regions are resized to 256×256.

We jointly combine various multi-modal object tracking benchmarks, including LasHeR \cite{lasher}, DepthTrack \cite{det}, and VisEvent \cite{visevent}, for the training process. 
UASTrack is evaluated on distributed multi-modal tasks across three RGB-T tracking benchmarks: LasHeR, RGBT234 \cite{RGBT234}, and GTOT \cite{GTOT}; one RGB-E benchmark: VisEvent; and one RGB-D benchmark: DepthTrack.

\subsection{Comparisons with State-of-the-art Approaches}

\textbf{RGB-D Tracking.} 
DepthTrack is a comprehensive RGB-D dataset comprising 150 training sequences and 50 test sequences, evaluated using F-score, Recall (Re), and Precision (Pr) metrics.
{UASTrack achieves F-score of 62.8\%,  Pr of 63.0\%, and Re of 62.5\% in Table \ref{tab1}.}
These results represent substantial improvements over the "Unified-All" tracker, Un-Track, with margins of 5.1\%, 3.7\%, and 7.4\% for F-score, Pr, and Re, respectively. 
Furthermore, UASTrack outperforms the "Unified-Architecture" tracker, SDSTrack, by 1.4\%, 1.6\%, and 1.1\% for the same metrics.

\textbf{RGB-T Tracking.}
LasHeR benchmark contains 979 training video sequences and 245 test video sequences, evaluated using three metrics: Precision Rate (PR), Success Rate (SR), and Normalized Precision Rate (NPR). 
{In Table \ref{tab1}, although the "Separated" models are better on LasHeR, UASTrack achieves strong and competitive results, surpassing the "Unified-Architecture" and "Unified-All" trackers, with SR, PR, and NPR of 57.0\%, 71.1\%, and 61.0\%, respectively, highlighting the effectiveness of the proposed unified framework.}

The RGBT234 benchmark integrating both RGB and thermal images includes a total of 234 test video sequences with nearly 116.7k frames. 
{As shown in Table \ref{tab2}, our UASTrack demonstrates competitive results, achieving SR of 65.1\% and PR of 87.6\%.}

The GTOT benchmark, which is designed to evaluate the robustness of RGB-T trackers, consists of 50 diverse video sequences. 
{As depicted in Table \ref{tab2}, UASTrack sets a new impressive performance with SR of 78.9\% and  PR of 93.3\%.
These results surpass the previous best-performing tracker, CKD, by margins of 1.7\% and 0.1\%, respectively.}

\textbf{RGB-E Tracking.} 
VisEvent consists of 500 video pairs for training and 320 video pairs for testing. 
{Table \ref{tab1} shows that UASTrack consistently outperforms VisEvent.
UASTrack attains the Precision Rate (PR) of 77.3\% and Success Rate (SR) of 61.0\%, surpassing best performing "Unified-All" tracker, M\textsuperscript{3}Track, by margins of 1.4\% and 0.6\%.}

\begin{figure}[t]
\centering
\includegraphics[width=0.5\textwidth]{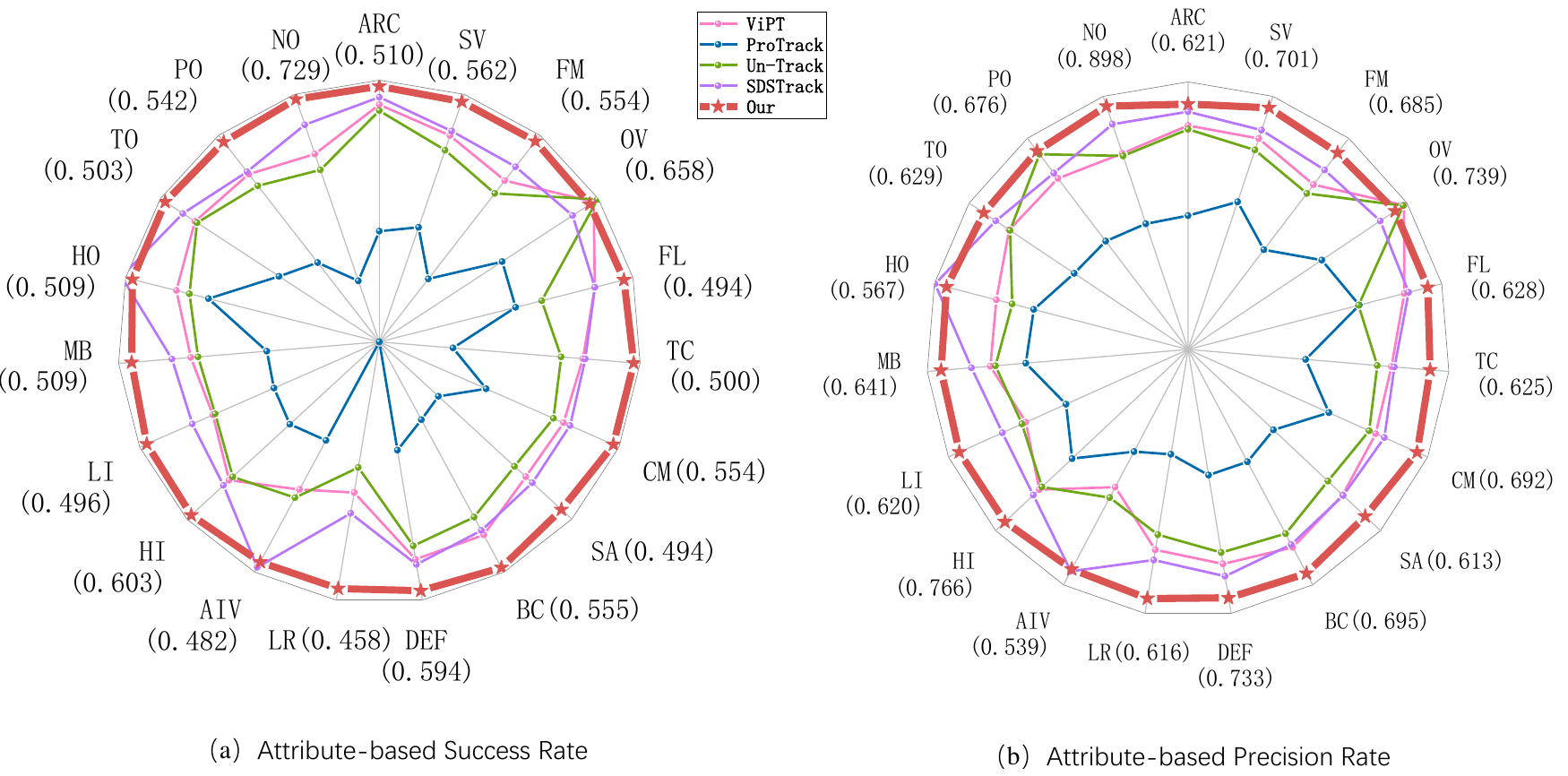}
\caption{The Success Rate (SR) and Precision Rate (PR) of 19 different attributes on the LasHeR dataset. (a) Attribute-based Success Rate (b) Attribute-based Precision Rate.}
\vspace{-3mm}
\label{fig:4}
\end{figure}

\textbf{Attribute-Based Performance on LasHeR.}
Our method is evaluated for various challenging attributes in comparison with existing SOTA trackers on the LasHeR dataset, as depicted in Fig. \ref{fig:4}. 
These attributes include No Occlusion (NO), Partial Occlusion (PO), Total Occlusion (TO), Hyaline Occlusion (HO), Motion Blur (MB), Low Illumination (LI), High Illumination (HI), Abrupt Illumination Variation (AIV), Low Resolution (LR), Deformation (DEF), Background Clutter (BC), Similar Appearance (SA), Camera Movement (CM), Thermal Crossover (TC), Frame Loss (FL), Out-of-View (OV), Fast Motion (FM), Scale Variation (SV), and Aspect Ratio Change (ARC). 
The experimental results show that our method achieves impressive performance compared with existing SOTA methods across most attributes in the metrics of SR and PR. 
Notably, it demonstrates a superior performance in scenarios involving DEF, FM, and SV, where the target objects experience drastic changes or blurring. 
Additionally, our UASTrack exhibits exceptional robustness in occlusion scenarios (HO, PO, and TO), addressing complex occlusion challenges very effectively.
Even under illumination changing environments, such as LI, HI, AIV, and TC, our UASTrack achieves a significantly improved tracking accuracy.

\subsection{Ablation Study}


\begin{table}[]
\centering
\caption{An ablation study of different variants of our model {on LasHeR, VisEvent, and DepthTrack. "UASTrack (FFT)" represents our proposed fully fine-tuned model. "UASTrack (Fixed)" denotes a fixed-structure unified model in which the MSA and TCA modules are replaced by a single shared branch, omitting modality-specific adapters.}}\vspace{0.3mm}
\renewcommand{\arraystretch}{1.4}
\scalebox{1}{
\label{tab_varients}
\begin{tabular}{cccccccc}
\hline
\multirow{2}{*}{Method} & \multicolumn{1}{c}{LasHeR} &\multicolumn{1}{c}{{VisEvent}} & \multicolumn{1}{c}{{DepthTrak}} \\
& SR 
& {SR} &{F-score} \\  \hline \hline
baseline & 0.482  & 0.588 &0.579
\\
MSA w/o sub-adapters & 0.561 
&0.603  &0.601    
\\
TCA w/o sub-adapters &0.554  &  0.557
& 0.556 \\
VA \& MSA share params &0.560
&0.602 &0.621
\\
{DAS with 100\% PSR} &0.564 &0.597 &0.619 \\
{UASTrack (FFT)} &0.554 &0.581 &0.528
\\
{UASTrack (Fixed)} &0.558 &0.591 &0.604
\\ 
{UASTrack (MOE)} & {0.557} &{0.592} &{0.601} \\
UASTrack &\textbf{0.570} &\textbf{0.610} &\textbf{0.628}
\\ \hline 
\end{tabular}}
\end{table}


\begin{figure}[t]
\centering
\includegraphics[width=0.45\textwidth]{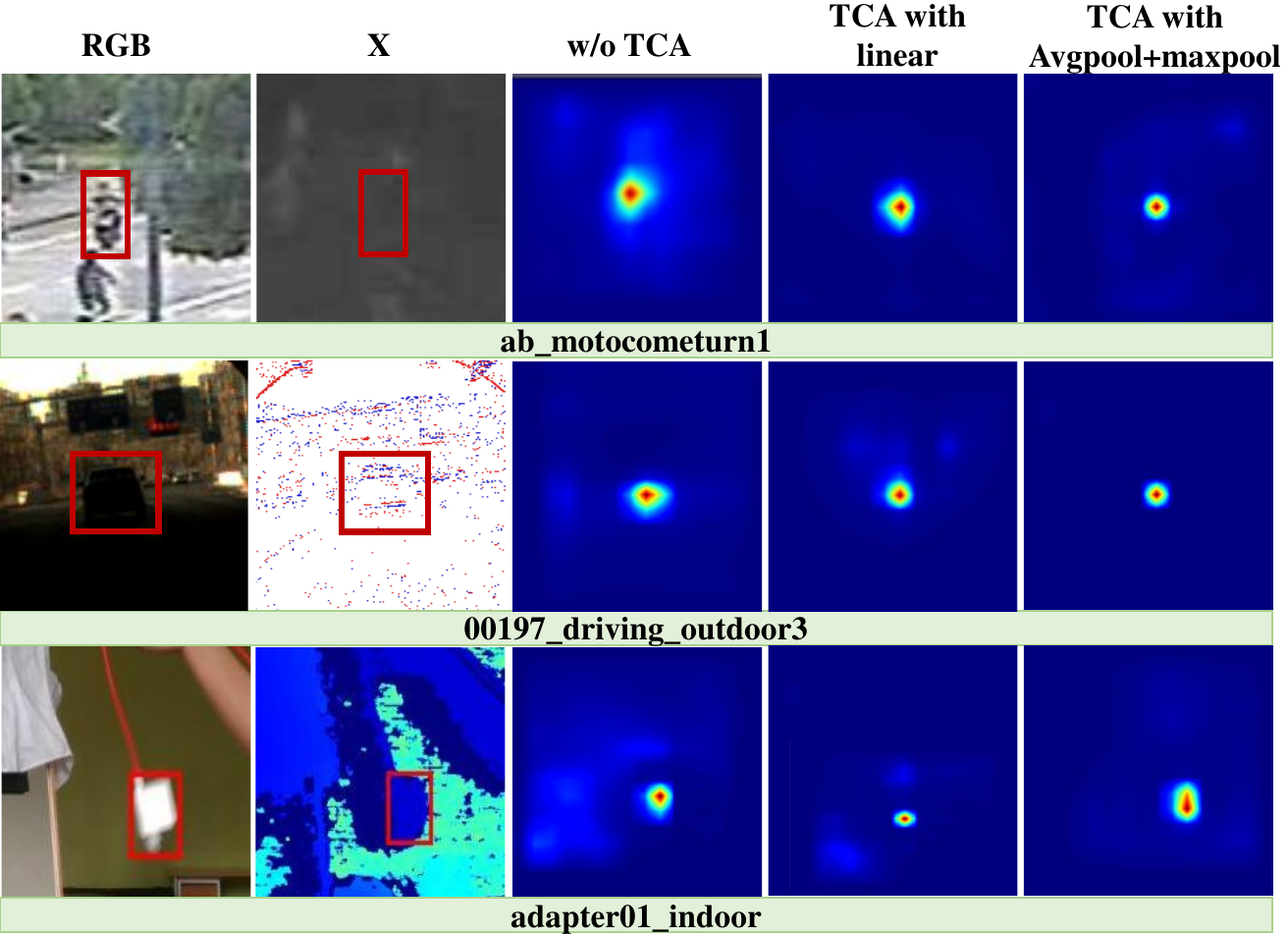}
\caption{Illustration of the score maps. "w/o TCA" represents UASTrack without the TCA module; "TCA with linear" represents that the TCA module exclusively employs linear layers; and "TCA with AvgPool+MaxPool" signifies that the TCA module integrates both average pooling and max pooling operations. 
}
\vspace{-3mm}
\label{fig:5}
\end{figure}

\textbf{An Ablation Study of Our Proposed Components}.
We conduct an ablation experiment to evaluate the different designs of our proposed UASTrack on the VisEvent, LasHeR, and DepthTrack benchmarks, as depicted in Table~\ref{tab3}. 
Since both the Classification Loss (CL) and the Task-Customized Adapter (TCA) rely on the modality type prediction carried out by the Discriminative Auto-Switch (DAS) module, their validation results for the individual modules are assessed based on the DAS module.
Firstly, the incorporation of DAS results in a significant improvement, with a 2.07\% increase for $\sigma$ compared to the baseline (first row). 
To be specific, the F-score on DepthTrack increases by 0.5\%, the SR on LasHeR improves by 5.2\%, and the SR on VisEvent rises by 0.3\%.
Even without CL (second row), the network demonstrates a superior performance in distinguishing the thermal modality compared to depth and event modalities.
This finding suggests that the thermal modality has inherent characteristics that make it more easily distinguishable by the network relative to the other modalities.
Integrating CL further enhances the network's performance (third row), leading to notable improvements, including a 1.6\% increase in SR on LasHeR, a 1.1\% rise in SR on VisEvent, and a 2.5\% boost in F-score on DepthTrack.
Additionally, the incorporation of TCA module improves the tracking accuracy (third row), contributing to a 1.2\% boost in SR on LasHeR, a 1.3\% increase in F-score on DepthTrack, and a 0.3\% improvement in SR on VisEvent.
When DAS, CL, and TCA are integrated into the foundation network, the optimal performance is achieved, with the SR of 57.0\% on LasHeR, the SR of 61.0\% on VisEvent, and the F-score of 62.8\% on DepthTrack.
{Introducing the DAS and TCA modules adds only minimal overhead, with parameter increases of 0.59M and 0.89M, and just 0.08G more FLOPs in total. 
This demonstrates the lightweight design of our proposed modules. 
Table ~\ref{tab3} shows that our method achieves consistent and significant performance improvements across multiple tasks, indicating that our design effectively balances efficiency and accuracy.}

\textbf{A Component Analysis of the Task-Customized Adapter.} 
We conduct an ablation experiment on the Task-Customized Adapter module across different tasks to investigate how the differences in the modality characteristics affect the adaptability of the network structures. 
As presented in Table \ref{tab4}, the results reveal that distinct modalities benefit from the tailored optimization strategies. 
For RGB-T tracking, employing a general adapter comprising linear layers achieves superior performance.
This approach results in a 1.4\% higher SR on LasHeR compared to the "w/ avgpool+maxpool" configuration.
In contrast, the depth and event data exhibit greater feature redundancy, which can introduce noise and hinder feature fusion.
To address this, integrating max pooling and average pooling operations into their respective sub-adapters enhances the TCA module's capacity.
{Using dual-pooling for the RGB-D and RGB-E tasks yields noticeable improvements over "w/ linear", with a 1.9\% gain in F-score on DepthTrack and a 0.5\% increase in SR on VisEvent.}
These findings highlight the necessity of customizing network structures to the unique characteristics of each modality, demonstrating that a one-size-fits-all approach is suboptimal for various multi-modal tracking tasks.

\textbf{The Inﬂuence of Different Adapter Variants and Our Model Variants.}
We conduct an ablation study to evaluate the effectiveness of different adapter variants on various tasks, as presented in Table~\ref{tab_varients}.
Three configurations are tested: (1) MSA without sub-adapters, where the MSA module uses a single shared adapter across all tasks instead of task-specific non-shared adapters; (2) TCA without sub-adapters, where the TCA module is similar to MSA without the sub-adapters and does not employ the modality-specific customization; and (3) VA and MSA with a shared adapter, replacing the original requirement of doubling the set of parameters for RGB and auxiliary features.
{When MSA does not incorporate sub-adapters, its performance drops significantly compared to UASTrack, highlighting the importance of using non-shared adapters for event, depth, and thermal modalities.
The performance of "TCA w/o sub-adapters" drops by 1.6\% in the SR of LasHeR, 5.3\% in the SR of VisEvent, and 7.2\% in the F-score of DepthTrack compared to our proposed UASTrack, further emphasizing the necessity of modality customization.
Furthermore, sharing parameters between the VA and MSA modules leads to performance drops of 1.0\% in SR on LasHeR, 0.8\% in SR on VisEvent, and 0.7\% in F-score on DepthTrack. These results suggest that applying a shared adapter to both the RGB and auxiliary modalities compromises the model’s ability to capture complementary information.}
Consequently, the design of non-shared parameters allows a better adaptation to the unique characteristics of each modality, effectively reducing information conflicts and improving the performance and generalization capability of multi-modal tasks.

We also conduct an ablation study to evaluate the effectiveness of our model variants.
{We compare the fully fine-tuned model of our UASTrack, denoted as UASTrack (FFT). 
The SR for the LasHeR dataset, the SR for the VisEvent dataset, and the F-score for the DepthTrack dataset decrease by 1.6\%, 11.9\%, and 10.0\%, respectively. 
Although increasing the number of trainable parameters from 2.10M to 94.24M ( Table \ref{tab7}) generally expands model capacity, in our setting, it leads to performance degradation due to several factors. 
Fully updating all parameters on relatively small multi-modal datasets causes catastrophic forgetting of the RGB feature space learned during pre-training. 
This weakens generalization across modalities, especially for event and depth data that differ greatly from the pre-training distribution. 
In contrast, our adapter-based fine-tuning updates only a small set of modality-specific parameters while keeping the backbone frozen, preserving the pre-trained representation and enabling more efficient adaptation for higher overall performance.
}
\begin{table}[]
\centering
\caption{An ablation study of the impact of parameter $\alpha$ {on LasHeR, VisEvent, and DepthTrack.}}\vspace{0.3mm}
\renewcommand{\arraystretch}{1.4}
\scalebox{1.1}{
\begin{tabular}{cccccc}
\hline
\multirow{2}{*}{$\alpha$} & \multicolumn{1}{c}{LasHeR} & \multicolumn{1}{c}{{VisEvent}} & \multicolumn{1}{c}{{DepthTrak}}  \\
& SR 
&{SR}  &{F-score}  \\  \hline  \hline
0.01 & 0.555 & 0.550 & 0.525 
\\
0.05 &0.564 & 0.603 & 0.611   
\\
0.1 &\textbf{0.570} &\textbf{0.610} &\textbf{0.628}
\\
0.5 & 0.565 &0.604 &0.609  
\\
1 & 0.565 & 0.602 & 0.603
\\
5 & 0.560 & 0.595 & 0.603
\\
10 & 0.553 &0.561 &  0.528 
\\ \hline
\end{tabular}}
\label{tab6}
\end{table}

\begin{table}[]
\centering
\caption{An ablation study of the channel low-rank dimensionality {on LasHeR, VisEvent, and DepthTrack. }}\vspace{0.3mm}
\renewcommand{\arraystretch}{1.4}
\label{c}
\scalebox{1.1}{
\begin{tabular}{cccccc}
\multicolumn{6}{c}{(a) VA \& MSA} \\ \hline
&  & 4 & {8} & 16 & 192 \\ \hline \hline
 LasHeR & SR & 0.552 & \textbf{0.570} & 0.556 & 0.557 \\
{VisEvent} & SR &0.550 &\textbf{0.610} &0.564&0.554 \\
{DepthTrack} 
&F-score &0.523 &\textbf{0.628}& 0.552 &0.525  
\\ 
\hline
\multicolumn{6}{c}{(b) TCA} \\ \hline 
&  & 8 & 96 & {\textbf{192}} & 384 \\ \hline \hline 
LasHeR& SR & 0.541 & 0.555 & \textbf{0.570} & 0.550 \\
{VisEvent} & SR &0.578 & 0.584 &\textbf{0.610} & 0.584 \\
{DepthTrack} 
& F-score  & 0.584 &0.598 &\textbf{0.628} &0.599 \\
\hline
\end{tabular}}
\label{tab8}
\end{table}
To further validate the core contribution of our adaptive multi-modal framework, we conduct an experiment by replacing the Modality-Specific Adapter (MSA) and Task Customization Adapter (TCA) modules with a fixed single-branch model that does not employ an adaptive mechanism. 
This UASTrack (Fixed) employs a shared backbone across all modalities without distinguishing between thermal, event, and depth features.
The UASTrack (Fixed) exhibits a consistent performance decline across all benchmarks compared to UASTrack. Specifically, on LasHeR, SR drops from 57.0\% to 55.8\% and PR from 71.1\% to 69.3\%; on VisEvent, SR decreases from 61.0\% to 59.1\% and PR from 77.3\% to 75.7\%; and on DepthTrack, the F-score declines from 62.8\% to 62.4\%.
These results clearly demonstrate the effectiveness and necessity of incorporating task-specific adaptations within our framework.

{We evaluate UASTrack with an MOE-based routing mechanism.
Specifically, we replace the MSA and TCA modules with an MoE structure containing 4 experts, select the top-2 experts, and remove the DAS module, forming UASTrack (MoE). The results show that UASTrack (MOE) achieves 0.557, 0.592, and 0.601 on LasHeR, VisEvent, and DepthTrak, respectively, which are slightly lower than our original UASTrack. 
This suggests that the simple hard routing mechanism is sufficiently effective for our task. Furthermore, we note that EMTrack also employs an MOE-based unified tracker as shown in Table I, but its performance remains lower than UASTrack, which further supports the effectiveness of our design choice.}
\begin{figure*}[t]
\centering
\includegraphics[width=0.98\textwidth]{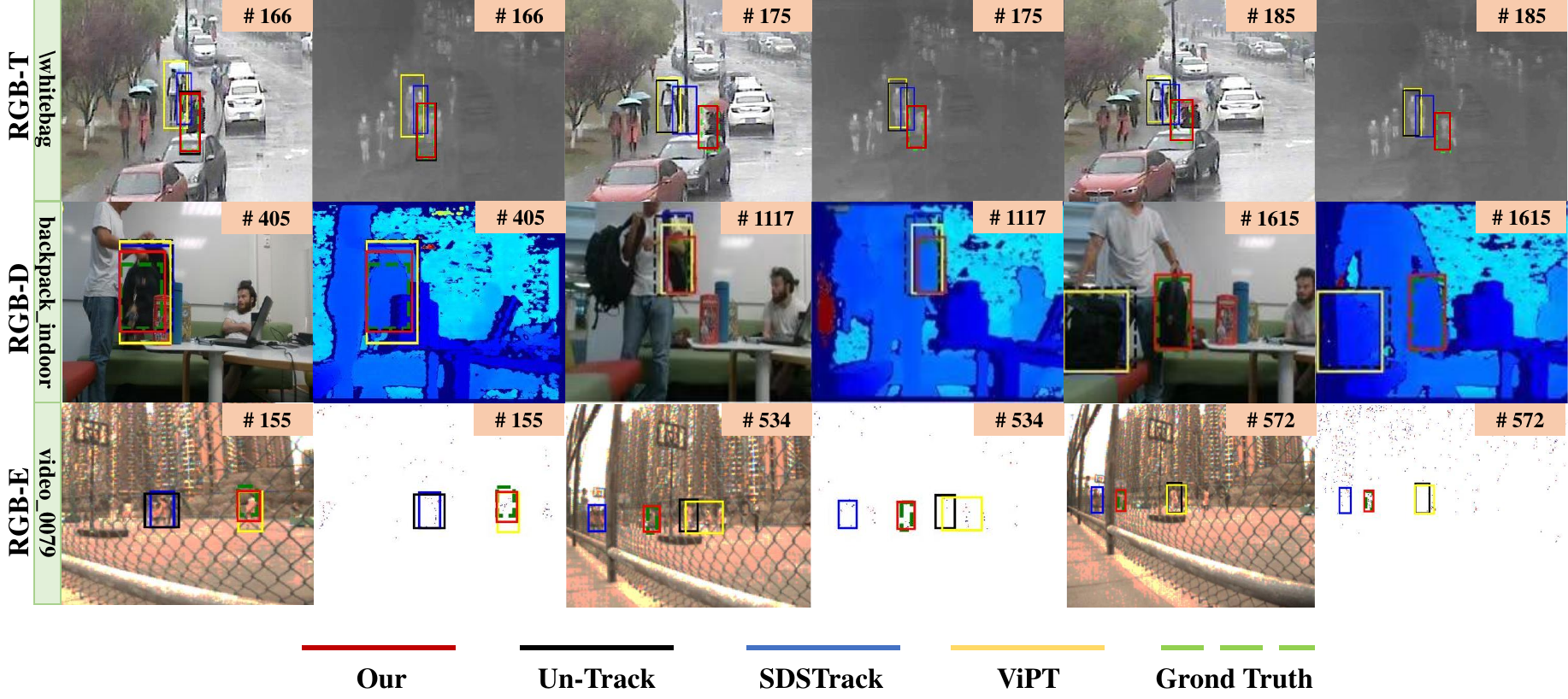}
\caption{A tracking results comparison. From top to bottom, we show the results on three video sequences, "\textit{whitebag}" from LasHeR dataset, "\textit{backpack\_indoor}" from DepthTrack dataset, and "\textit{video\_0079}" from VisEvent dataset.}
\vspace{-3mm}
\label{fig:6}
\end{figure*}

\textbf{The Inﬂuence of Parameter $\alpha$.}
The selection of hyperparameters is crucial for optimizing the object tracking performance.
We explore the effect of parameter $\alpha$, while keeping $L_{1}$ and $L_{GIoU}$ consistent with the OSTrack baseline \cite{ostrack}.
{As shown in Table~\ref{tab6}, we use $\alpha = 1$ as a reference point and explore values ranging from 1/100 to 10 times this value, specifically selecting 0.01, 0.05, 0.1, 0.5, 1, 5, and 10. To maintain balance across the range, we use 0.05, 0.5, and 5 as median candidates from the lower and upper intervals.
The results indicate that setting $\alpha = 0.1$ yields the best performance across all three datasets, achieving the highest SR on LasHeR (0.570), SR on VisEvent (0.610), and F-score on DepthTrack (0.628). Compared to smaller values such as $\alpha = 0.01$ and larger ones like $\alpha = 1$ and $\alpha = 10$, $\alpha = 0.1$ leads to improvements of up to 0.5\%–10.5\% depending on the dataset.
These results suggest that a smaller $\alpha$ may assign an insufficient weight to the critical components while a larger $\alpha$ may overemphasize certain aspects.}

\begin{table}[]
\centering
\caption{A comparison of the computational costs and the speed in Frames Per Second (FPS) of the different trackers on LasHeR test set. "*" indicates the results reproduced by us using unified training on the combined LasHeR, VisEvent, and DepthTrack datasets for fair comparison. {Values in '()' denote the training flops or training parameters, while those outside '()'indicate the total model parameters.}}\vspace{0.3mm} 
\renewcommand{\arraystretch}{1.5}
\scalebox{1}{
\begin{tabular}{ccccc}
\hline
\multirow{1}{*}{Method} & \multirow{1}{*}{{Params(M)}} & \multirow{1}{*}{{Flops(G)}} & \multirow{1}{*}{FPS} & \multirow{1}{*}{SR} \\ \hline \hline 
OSTrack-RGBT &92.13 &56.44 &71.28 & 0.479    
\\
OSTrack-RGB  &92.13 & 29.24 &107.10&0.470
\\
OSTrack-TIR &92.13 &29.24  &107.65&0.453
\\ \hline
ViPT & 92.97 (0.84) & 59.56 (3.12) & - & 0.525 
\\
SDSTrack & - & - & 20.90 & 0.531  
\\
TBSI & 191.36 & 79.80 & 32.00 & 0.556  
\\ 
Un-Track & 98.78 (6.65) & 58.58 (2.14) & - & 0.536 
\\ 
{EMTrack} & 108.13 (16.00) &58.44 (2.00) &34.00 &0.533 \\
{SUTrack-B224 *} & 155.00 
& 23.00
& 34.58
& 0.563 \\ 
{UASTrack-FFT} & 94.24
& 58.40
& 37.29 & {0.554} \\
UASTrack & 94.24 (2.10)
& 58.39 (1.95) 
& 43.93 & \textbf{0.570} \\ \hline
\end{tabular}}
\label{tab7}
\end{table}

\textbf{A Low-Rank Dimensionality Analysis.}
We evaluate the effectiveness of different channel low-rank dimensionalities for the Visual Adapter (VA), Modality-Specific Adapter (MSA), and Task-Customized Adapter (TCA) as presented in Table \ref{tab8}.
{In our work, the term “low-rank” refers to the reduction of the channel dimension. 
We use low-rank approximation to reduce the number of channels in the feature representation. 
This helps in improving computational efficiency while maintaining performance. }
{Since both VA and MSA are applied during the feature extraction process, while TCA operates after the feature extraction, VA and MSA use the same low-rank dimensionality for the channel.
We experiment by varying the ranks of VA and MSA across four configurations: 4, 8, 16, and 192, as shown in Table \ref{tab8} (a). 
A rank of 8 yields the best overall performance, achieving an SR of 57.0\% on LasHeR, 61.0\% on VisEvent, and F-score of 62.8\% on DepthTrack. 
In contrast, setting the dimensionality to 4 results in degraded performance, with SR values decreasing to 55.2\% on LasHeR, 55.0\% on VisEvent, and the F-score dropping to 52.3\% on DepthTrack. Increasing the dimensionality to 192 also leads to suboptimal results, with SR of 55.7\% on LasHeR and 55.4\% on VisEvent, and F-score of 52.5\% on DepthTrack.
The results reveal that both very low and very high ranks degrade performance.}

Similarly, as shown in Table \ref{tab8} (b), we examine TCA across the rank configurations of 8, 96, 192, and 384. Our findings demonstrate that a rank of 192 yields the best performance.
In summary, during the feature extraction stage, the adapter dimensionality should be maintained at a relatively low value to minimize computational overhead and improve performance.
At the task-level stage, the dimensionality should be set higher to enhance feature representation capabilities and accommodate task complexity requirements.

\subsection{Qualitative Evaluation}
\textbf{A Qualitative Analysis of the  Task-Customized Adapter.}
To evaluate the effectiveness of the Task-Customized Adapter in achieving modality-specific customization, we conduct a qualitative analysis using selected sequences from three tasks. 
Specifically, we analyze the sequence "\textit{ab\_motocometurn1}" from the LasHeR dataset, the sequence "\textit{00197\_driving\_outdoor3}" from the VisEvent dataset, and the sequence "\textit{adapter01\_indoor}" from the DepthTrack dataset, as shown in Fig. \ref{fig:5}.
The max pooling \cite{Inceptiontransformer, DropanOctave} emphasizes the prominent responses in sparse signals, such as locally active areas in the event and depth streams, by retaining the most significant local features. 
In contrast, average pooling \cite{bullier2001integrated,kauffmann2014neural} calculates regional averages and reduces redundancy, making it well-suited for processing local geometric and fast motion information. 
Combining these two pooling operations complements their strengths, enabling the extraction of local and global information.
As illustrated in Fig. \ref{fig:5}, the TCA module optimizes the multi-modal features very effectively, whether inserted in all linear layers or enhanced with average and max pooling.
For the sequence "\textit{ab\_motocometurn1}", the combination of RGB-T features provides richer target information compared to the sparse characteristics of depth and event data.
Consequently, the TCA module with linear layers has enough capacity for RGB-T tracking.
{Sequences "\textit{00197\_driving\_outdoor3}" and "\textit{adapter01\_indoor}", demonstrate that dual-pooling strategy in TCA preserves critical features such as geometric edges in RGB-D tracking and motion-related blobs in RGB-E tracking, while effectively filtering out background noise.}
The visual analysis of different TCA configurations across the various tasks further confirms that employing modality-specific adapters for thermal, depth, and event data improves the adaptation of multi-modal features to the RGB-based pre-trained embedding space.

\begin{table}[]
\centering
\caption{{Fair Comparison of deployment settings and performance between SUTrack and the proposed UASTrack. The last line "SR" is the metric on LasHeR. "*" indicates the results reproduced by us using unified training on the combined LasHeR, VisEvent, and DepthTrack datasets for fair comparison.}}\vspace{0.3mm}
\renewcommand{\arraystretch}{1.3}
\label{deploymentmetric}
\scalebox{0.95}{
\begin{tabular}{ccccc}
\hline
\multicolumn{2}{c}{Tracker} & \begin{tabular}[c]{@{}c@{}}SUTrack\\ -B224\end{tabular} & \begin{tabular}[c]{@{}c@{}}SUTrack\\ -B224 *\end{tabular} & UASTrack \\ \hline \hline
\multirow{9}{*}{\begin{tabular}[c]{@{}c@{}}Traning \\ Dataset\end{tabular}} & COCO &\checkmark  &  &  \\
 & LaSOT & \checkmark &  &  \\
 & GOT10k & \checkmark &  &  \\
 & TrackingNet &\checkmark  &  &  \\
 & TNL2K &\checkmark  &  &  \\
 & VASTTrack & \checkmark &  &  \\
 & DepthTrack & \checkmark &\checkmark  & \checkmark \\
 & VisEvent & \checkmark & \checkmark & \checkmark \\
 & LasHeR &  \checkmark& \checkmark & \checkmark \\ \hline
\multirow{11}{*}{\begin{tabular}[c]{@{}c@{}}Deployment \\ metrics \\ and device\end{tabular}} & GPU &4×A400 & 1×3090Ti & 1×3090Ti \\
&Latency(ms) &-&638.13& 978.39 \\
 & Epochs & 180 & 180 & 80 \\
 & Transformer & HiViT-B & HiViT-B & \begin{tabular}[c]{@{}c@{}}OSTrack\\ -256\end{tabular} \\
 & \begin{tabular}[c]{@{}c@{}}Research \\ -resolution\end{tabular} & 224×224 & 224×224 & 256×256 \\
 & \begin{tabular}[c]{@{}c@{}}Template \\ -resolution\end{tabular} & 112×112 & 112×112 & 128×128 \\
 & Data augmentation   &Common &Common  &Common \\
&Frozen policy   &FFT &FFT & Adapter \\
 & Peak memory & - &{21.48GB}  & {12.87GB}\\
 & SR & \textcolor{red}{\textbf{0.599}} & 0.563 & \textcolor{blue}{\textbf{0.570}} \\ \hline
\end{tabular}}
\end{table}

\textbf{A Qualitative Analysis of Tracking Results on Three Challenging Scenarios.}
{To demonstrate the advantages of our proposed method in real tracking scenarios, we visualize the tracking results of UASTrack against state-of-the-art trackers on RGB-T, RGB-E, and RGB-D tracking tasks, as illustrated in Fig. \ref{fig:6}.}
{We carefully select three representative sequences from challenging scenarios. 
In the "\textit{whitebag}" sequence, the target is located in an environment with frequent thermal crossovers involving other hot objects, leading to degradation in the thermal modality. Additionally, the presence of rainfall reduces the quality of RGB images.
Despite these challenges, UASTrack demonstrates resilience to low-quality thermal and RGB images, consistently capturing the target’s precise location throughout the tracking process. 
In contrast, previous methods exhibit varying degrees of tracking drift.
In the "\textit{backpack\_indoor}" sequence, the target is a black backpack, which faces interference from another visually similar black backpack. 
At frame 405, severe occlusion further complicates accurate localization in RGB images.
Notably, UASTrack effectively leverages the complementary properties of these modalities by consistently generating precise bounding boxes.
The "\textit{video\_0079}" sequence presents two major challenges commonly encountered in visual object tracking: fast motion and severe occlusion. 
Among the compared methods, only UASTrack consistently and accurately localizes the target without being misled by distractions.
This observation strongly demonstrates that our unified approach successfully achieves adaptive task selection while enhancing the robust representation of multi-modal complementary features.}

\begin{table}[]
\centering
\caption{{Performance evaluation under modality missing and asynchronous modalities on DepthTrack, LasHeR, and VisEvent. "Missing X" denotes the absence of the auxiliary modality, while "Delay frames" indicates the frame delay applied to the auxiliary stream.}}\vspace{0.3mm}
\renewcommand{\arraystretch}{1.6}
\scalebox{1}{
\label{missX}
\begin{tabular}{ccccc}
\hline
\multicolumn{2}{c}{\multirow{2}{*}{}} & DepthTrack & LasHeR & VisEvent \\
\multicolumn{2}{c}{} & F-score & SR & SR \\ \hline \hline
Missing & RGB & 0.542 & 0.463 & 0.545 \\
modality & X (T,D,or E) & 0.589 & 0.499 & 0.594 \\ \hline
\multirow{3}{*}{\begin{tabular}[c]{@{}c@{}}Delay \\ frames\end{tabular}} & 1 & 0.596 & 0.545 & 0.592 \\
 & 3 & 0.579 & 0.511 & 0.581 \\
 & 5 & 0.571 & 0.491 & 0.577 \\ \hline
\end{tabular}}
\end{table}

\subsection{Explorative Analysis}
\begin{figure}[t]
\centering
\includegraphics[width=0.4\textwidth]{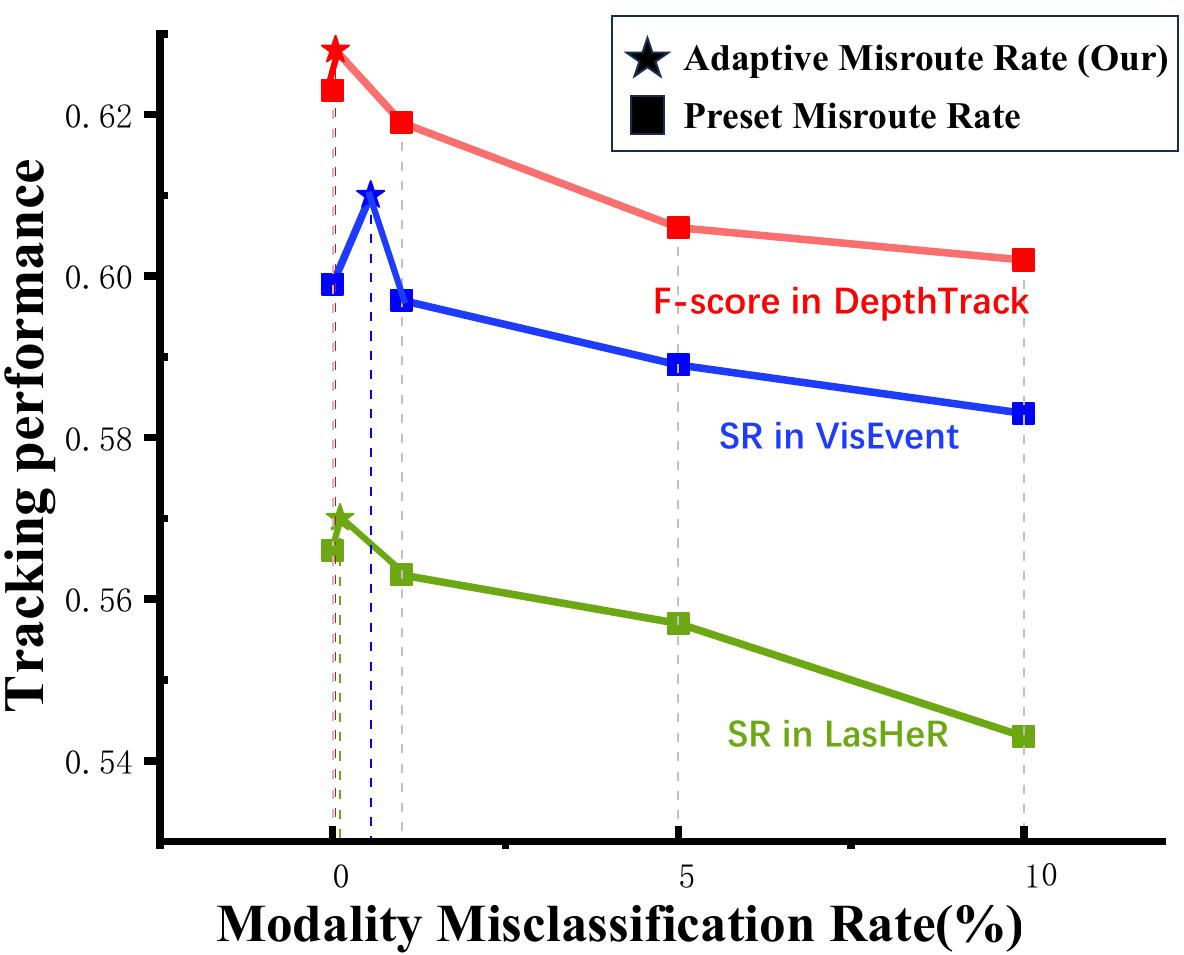}
\caption{{Exploration analysis of different modality misrouting rates on LasHeR, VisEvent, and DepthTrack.}}
\vspace{-3mm}
\label{fig:71}
\end{figure}

{\textbf{Robustness to misrouting.} 
As for modality misclassification, we conduct an additional quantitative experiment to analyze the effect of modality misclassification on the overall tracking accuracy. 
We simulate modality misclassification by introducing preset misrouting rates (under two-stage training) of 0\%, 1\%, 5\%, and 10\%, and by applying our end-to-end training strategy. The performance is evaluated on the LasHeR, VisEvent, and DepthTrack datasets.
For the preset misroute setting, we adopt a two-stage training scheme: we first train a Discriminative Auto-Selector (DAS) that predicts the corresponding misroute rate, and during the second stage, its parameters are frozen while training the remaining model.
The results are summarized in Fig. \ref{fig:71}.
For small misrouting rates (<1\%), the performance degradation is minimal compared to our training scheme, while for larger misrouting rates (>5\%), the performance drops significantly. 
Interestingly, when the Discriminative Auto-Selector achieves a 0\% misclassification rate, accurately predicting the types of auxiliary modality, the performance does not improve further. 
This occurs because auxiliary modalities share common information, and when accurate predictions are made, the Modality-Specific Adapter and Task-Customized Adapter modules' sub-adapters receive only the modality-specific information. 
As a result, they miss out on the shared information from other modalities, which would have enhanced performance within a small range of misclassification rates.
}

{\textbf{Missing/async modalities (hard routing).}
We evaluate the impact of missing or asynchronous modalities by either removing a modality or introducing temporal desynchronization (simulated by delaying the auxiliary stream), as shown in Table \ref{missX}. 
When the auxiliary modality is missing, the system defaults to the RGB-only branch, leading to performance degradation due to the lack of complementary information. 
When one modality is missing, we fill the missing stream by duplicating the available modality. Specifically, if the RGB modality is missing, the auxiliary input is copied into the RGB branch. If the auxiliary modality is missing, both branches share the VA modules, and the DAS and MSA modules are disabled accordingly. 
When either modality is removed, the performance declines due to reduced complementary information.
To simulate the impact of real-world temporal misalignments, we introduce delays in the auxiliary modality. Specifically, for each RGB frame at time $t$, the corresponding auxiliary modality frame is shifted to $t + \Delta$ with $\Delta = 1, 3, 5$ frames. As shown in Table \ref{missX}, the performance decreases with increasing delay.
For example, on DepthTrack, the F-score drops to 0.571 for $\Delta = 5$, demonstrating the negative impact of temporal desynchronization.}

\textbf{Cross-Modal Dependency Analysis.}
Although depth, event, and thermal modalities exhibit substantial differences, they inherently possess both shared and complementary information. This enables the analysis of cross-modality dependency and transferability by routing the input video data into the appropriate branches, that have been trained on different modalities, as illustrated in Table \ref{tab5}.
Depth demonstrates the lowest transferability to event and thermal modalities, with the largest negative changes observed: a decrease of -10.8\% in $\sigma$ when depth features are transferred to the thermal branch and -18.5\% when transferred to the event branch (Row \#4, \#5, \#6).
This indicates that the performance of the depth modality is significantly degraded when its features are utilized in other modalities.
Although the event and thermal modalities exhibit better cross-modality robustness, they still experience notable reductions in SR and PR.
The thermal modality is moderately transferable, with $\sigma$ reductions of -8.8\%, when transferred to the event branch, and -12.3\%, when transferred to the depth branch (Row \#1, \#2, \#3).
The event modality exhibits the highest transferability, with a $\sigma$ reduction of less than 7\%, highlighting its comparative resilience in cross-modality transfer (Row \#7, \#8, \#9).

These findings suggest that depth data has limited generalizability, likely attributable to its strong dependence on structural and geometric information.
In contrast, the event data generalizability is greater, potentially due to its sparse and dynamic characteristics, which afford more flexibility across diverse tasks.
Thermal features exhibit moderate transferability, occupying an intermediate position relative to depth and event data in terms of cross-modality robustness.

{\textbf{A Comparison of SUTrack and the Proposed UASTrack in terms of the Deployment Settings and their  Performance.}
Table IX summarizes the deployment metrics and experimental configurations for SUTrack and our proposed UASTrack. To ensure a fair comparison, we reproduce a variant of SUTrack, denoted as SUTrack-B224*, by retraining it under the same unified training protocol as UASTrack. Specifically, both methods are trained on the combined LasHeR, VisEvent, and DepthTrack datasets, using identical GPU hardware (1×3090Ti) and comparable transformer backbones (HiViT-B for SUTrack-B224* and OSTrack-256 for UASTrack). 
The input and template resolutions are also aligned to maintain consistent visual scales.
It is worth noting that SUTrack utilizes large-scale RGB tracking and RGB-language training datasets such as COCO, LaSOT, GOT10k, TrackingNet, TNL2K, and VASTTrack.
In contrast, UASTrack employs a dual-branch design, consisting of an RGB branch and an auxiliary branch. 
The stronger performance of SUTrack can therefore be partly attributed to its access to a much larger amount of pretraining data, enabling it to build a more powerful feature representation.}

\begin{table*}[]
\centering
\caption{An ablation experiment exploring the cross-modal dependency on different multi-modal tracking benchmarks. The \#1, \#4, and \#7 rows show the performance of our method on the RGB-T, RGB-D, and RGB-E tracking tasks, respectively. Type "X$\rightarrow$Y" indicates that the X modality features are fed into the Y branch during testing. 
The column $\sigma$ represents the average percentage change across all metrics compared to the original task.
}\vspace{0.3mm} 
\renewcommand{\arraystretch}{1.2}
\scalebox{1.2}{
\begin{tabular}{cccccccccc}
\hline
\multirow{2}{*}{Row}&\multirow{2}{*}{Type} & \multicolumn{2}{c}{LasHeR} & \multicolumn{2}{c}{VisEvent} & \multicolumn{3}{c}{DepthTrack} & \multirow{2}{*}{$\sigma$} \\
& & SR & PR & SR & PR & Pr & Re & F-score &  \\ \hline \hline
\#1 &Thermal & 0.564 & 0.703 & - & - & - & - & - & - \\
\#2 &Thermal-\textgreater{}Event & 0.481 & 0.610 & - & - & - & - & - & -8.8\% \\
\#3 &Thermal-\textgreater{}Depth & 0.439 & 0.581 & - & - & - & - & - & -12.3\% \\
\#4 &Depth & - & - & - & - & 0.630 & 0.625 & 0.628 & - \\
\#5 &Depth-\textgreater{}Thermal & - & - & - & - & 0.521 & 0.518 & 0.520 & -10.8\% \\
\#6 &Depth-\textgreater{}Event & - & - & - & - & 0.548	&0.551	&0.550  & -18.5\% \\
\#7 &Event & - & - & 0.607 & 0.773 & - & - & - & - \\
\#8&Event-\textgreater{}Thermal & - & - & 0.535 & 0.709 & - & - & - & -6.8\% \\
\#9 &Event-\textgreater{}Depth & - & - & 0.531 & 0.712 & - & - & - & -6.9\% \\ \hline
\end{tabular}}
\label{tab5}
\end{table*}

{For data augmentation, we adopt common strategies, consistent with SUTrack, including horizontal flipping and brightness jittering during training. Unlike SUTrack, which applies "FFT" (full fine-tuning), our approach utilizes adapter learning, which decreases the computational burden.
This controlled setup effectively removes discrepancies in data volume and resolution that are inherent to the original SUTrack configuration, leading to a fair comparison. 
As shown in Table IX, under these fair and unified settings, UASTrack requires only 12.87 GB of peak memory compared with 21.48 GB for SUTrack-B224*, while achieving a higher SR of 0.570, demonstrating both superior computational efficiency and competitive accuracy.
}

\begin{table}[]
\centering
\caption{{Classification Accuracy Analysis of DAS Module on LasHeR, VisEvent, and DepthTrack.}}
\vspace{0.5mm} 
\renewcommand{\arraystretch}{1.4}
\scalebox{1}{
\begin{tabular}{cccc}
\hline
Input Type & {Fail / Total} & {Accuracy} & {Misclassification} \\ \hline \hline
T & 241 / 220703 & 99.89\% & T$\rightarrow$E (231 times) \\
D & 337 / 80410 & 99.58\% & D$\rightarrow$E (337 times) \\
E & 869/ 157799 & 99.45\% & E$\rightarrow$D (841 times) \\ \hline
\end{tabular}}
\label{tab:6}
\end{table}

\begin{figure}[t]
\centering
\includegraphics[width=0.45\textwidth]{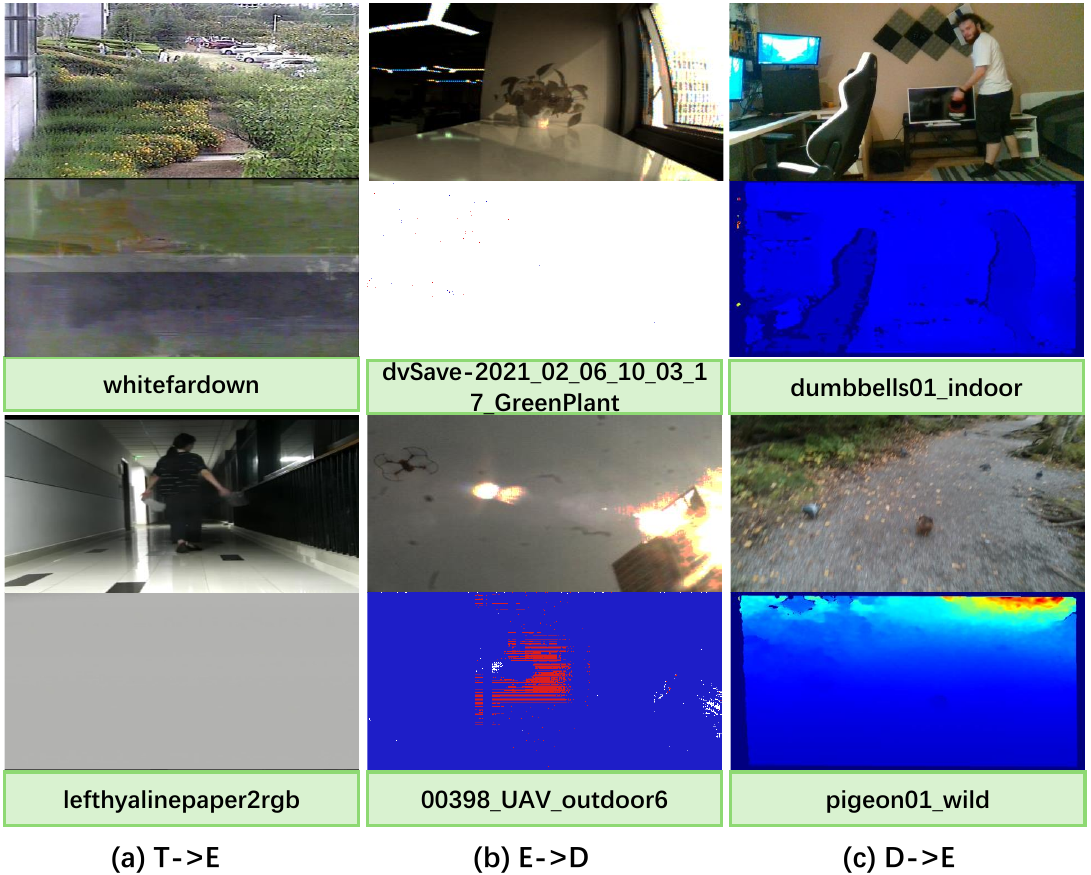}
\caption{{Illustration of the most common misclassifications of DAS module.
From left to right, we show the misclassification images on six video sequences, "\textit{whitefardown
}" and "\textit{lefthyalinepaper2rgb}" from LasHeR dataset, "\textit{dvSave\-2021\_02\_06\_10\_03\_17\_GreenPlant}" and "\textit{00398\_UAV\_outdoor6}" from VisEvent dataset, and "\textit{dumbbells01\_indoor}" and "\textit{pigeon01\_wild}" from DepthTrack dataset.}}
\vspace{-3mm}
\label{fig:8}
\end{figure}

{\textbf{A Comparison of the Computational Costs and Speed.} 
Table~\ref{tab7} presents a comparison of our proposed UASTrack with SOTA trackers in terms of tracking speed, parameters, flops, and overall performance.
The comparison includes both separately trained models, such as TBSI, and unified models, including ViPT, SDSTrack, Un-Track, and SUTrack.
We compare our model with SUTrack-B224 because our model uses OSTrack-256 as the baseline, and their model sizes are similar. To ensure a fair comparison, we use the combination of training data from DepthTrack, VisEvent, and LasHeR to train the SUTrack-B224.
UASTrack adopts adapter-based fine-tuning for multi-modal object tracking, substantially reducing the number of trainable parameters and flops compared to fully fine-tuned methods such as TBSI and SUTrack.
UASTrack requires only 2.10M training parameters and 1.95G training flops, saving up to 98\% of storage space compared with the full fine-tuning model of SUTrack.
UASTrack achieves an inference speed of 43.93 FPS, outperforming SDSTrack, TBSI, and SUTrack-B224 by 23.03, 11.93, 13.79, and 6.64 FPS, respectively.
In addition, UASTrack delivers superior accuracy, achieving SR and PR of 57.0\% and 71.1\%. 
Compared to SUTrack-B224, the “Unified-All” tracker, UASTrack achieves improvements in tracking accuracy, speed, and computational efficiency.}


{\textbf{An Analysis of Classification Accuracy and Misclassifications of DAS Module.}
Table \ref{tab:6} presents the classification accuracy and misclassification trends of the DAS module across three modality types. 
The model achieves high accuracy for all modalities, with thermal being the most accurately identified (99.89\%), followed by Depth (99.58\%) and Event (99.45\%).
Specifically, thermal and depth are occasionally misclassified as event, while event is frequently confused with depth, accounting for 841 misclassifications.}

{Additionally, we perform a visual analysis of the most common misclassifications. 
As illustrated in Fig. \ref{fig:8}, the DAS module tends to misclassify inputs when the quality of the input modality is particularly low. 
Sequences “\textit{whitefardown}” and “\textit{lefthyalinepaper2rgb}” are thermal modality inputs of low quality that are misclassified as event data. 
Sequences “\textit{dvSave\-2021\_02\_06\_10\_03\_17\_GreenPlant}” and “\textit{00398\_UAV\_outdoor6}” contain degraded event inputs that are predicted as depth data, while “\textit{dumbbells01\_indoor}” and “\textit{pigeon01\_wild}” involve low-quality depth inputs that are misclassified as event data.
In such cases, the feature distributions of depth, thermal, and event modalities become similar, increasing the likelihood of cross-modal misclassification.}

\section{Conclusion}
In this paper, we present a novel unified multi-modal tracker that incorporates modality-based customization and adaptive processing for application to multi-modal object tracking. 
Specifically, we propose a Discriminative Auto-Switch to enable a dynamic adaptation of the multi-modal tracker to different tasks.
Additionally, we introduce a Task Customization Adapter to facilitate task-specific customization of the video information flow through the tracker architecture, thereby enhancing the robustness and accuracy of tracking.
Our approach not only mitigates the discrepancy between a single-modality pre-training and a general multi-modal deployment,  but also establishes a unified multi-modal tracker capable of operating without the prior knowledge of the input modality types. 
The experimental results demonstrate the effectiveness of our method, showing significant improvements in all RGB-X tracking scenarios.
While our method achieves competitive performance across multiple tasks, it still faces challenges under adverse weather conditions such as rain, fog, and low illumination. 
In future work, we will extend our framework in adverse weather and design adaptive fusion strategies to ensure robust perception in degraded visual environments.

\section*{Acknowledgments}
This work is supported in part by the National Key Research and Development Program of China (2023YFF1105102, 2023YFF1105105), the National Natural Science Foundation of China (Grant NO. 62020106012, 62332008, 62106089, U1836218, 62336004) and the 111 Project of Ministry of Education of China (Grant No.B12018).





\bibliographystyle{IEEEtran}
\bibliography{ref}


\begin{IEEEbiographynophoto}{He Wang} is now a Ph.D. student with the School of Artificial Intelligence and Computer Science, Jiangnan University. Her research interests include multi-modal object tracking and deep learning.\end{IEEEbiographynophoto}

\begin{IEEEbiographynophoto}{Tianyang Xu} (Member, IEEE) received the B.Sc. degree in electronic science and engineering from Nanjing University, Nanjing, China, in 2011, and the
Ph.D. degree from the School of Artificial Intelligence and Computer Science, Jiangnan University, Wuxi, China, in 2019. He is currently an Associate Professor with the School of Artificial Intelligence and Computer Science, Jiangnan University. His research interests include visual tracking and deep learning.
\end{IEEEbiographynophoto}

\begin{IEEEbiographynophoto}{Zhangyong Tang} is now a Ph.D. student with the School of Internet of Things Engineering, Jiangnan University. His research interests include multi-modal object tracking and deep learning.\end{IEEEbiographynophoto}

\begin{IEEEbiographynophoto}{Xiao-Jun Wu} received the B.Sc. degree in mathematics from Nanjing Normal University, Nanjing, China, in 1991, and the M.S. and Ph.D. degrees in pattern recognition and intelligent systems from Nanjing University of Science and Technology, Nanjing, in 1996 and 2002, respectively. He is currently a Professor in artificial intelligence and pattern recognition with Jiangnan University, Wuxi, China. His research interests include pattern recognition, computer vision, fuzzy systems, neural networks, and intelligent systems. He is currently a fellow of IAPR and AAIA. He has won several domestic
and international awards because of his research achievements. He served as an associate editor for several international journals.\end{IEEEbiographynophoto}

\begin{IEEEbiographynophoto}{Josef Kittler} received the B.A., Ph.D., and D.Sc. degrees from the University of Cambridge, in 1971, 1974, and 1991, respectively. He is a distinguished Professor of Machine Intelligence at the Centre for Vision, Speech and Signal Processing, University of Surrey, Guildford, U.K. He conducts research in biometrics, video and image database retrieval, medical image analysis, and cognitive vision. He published the textbook Pattern Recognition: A Statistical Approach and about 1000 scientific papers. His publications have been cited by around 70,000 times.

He is series editor of Springer Lecture Notes on Computer Science. He currently serves on the Editorial Boards of Pattern Recognition Letters, Pattern Recognition and Artificial Intelligence, Pattern Analysis and Applications. He also served as a member of the Editorial Board of IEEE Transactions on Pattern Analysis and Machine Intelligence during 1982-1985. He served on the Governing Board of the International Association for Pattern Recognition (IAPR) as one of the two British representatives during 1982-2005, and the President of IAPR during 1994-1996.\end{IEEEbiographynophoto}

\newpage

 





\end{document}